\documentclass[10pt,twocolumn,letterpaper]{article}

\usepackage[pagenumbers]{cvpr} 

\usepackage[dvipsnames]{xcolor}

\usepackage{kotex}
\usepackage{lipsum}
\usepackage{pifont}
\usepackage{mathtools}

\usepackage{array}
\newcolumntype{P}[1]{>{\centering\arraybackslash}p{#1}}
\newcolumntype{M}[1]{>{\centering\arraybackslash}m{#1}}
\newcolumntype{L}[1]{>{\hspace{0.5em}\raggedright\arraybackslash}m{#1}}
\newcolumntype{R}[1]{>{\raggedleft\arraybackslash}m{#1}}

\usepackage{xspace}

\makeatletter
\DeclareRobustCommand\onedot{\futurelet\@let@token\@onedot}
\def\@onedot{\ifx\@let@token.\else.\null\fi\xspace}

\makeatother

\usepackage{caption}
\usepackage{tabularx,booktabs}
\newcolumntype{Y}{>{\centering\arraybackslash}X}
\usepackage{multirow}
\usepackage{soul}
\newcommand{\drule}{\specialrule{0.2pt}{1pt}{1pt}%
            \specialrule{0.2pt}{0pt}{\belowrulesep}%
            }

\usepackage{algorithm,algpseudocode}
\usepackage[algo2e]{algorithm2e} 
\definecolor{commentcolor}{RGB}{110,154,155}
\definecolor{defcolor}{RGB}{225,81,145}

\definecolor{cvprblue}{rgb}{0.21,0.49,0.74}
\usepackage[pagebackref,breaklinks,colorlinks,allcolors=cvprblue]{hyperref}

\def\paperID{3508} 
\def\confName{CVPR}
\def\confYear{2026}

\title{EgoX: Egocentric Video Generation from a Single Exocentric Video}

\author{
    Taewoong Kang$^{1*}$,
    Kinam Kim$^{1*}$,
    Dohyeon Kim$^{2*}$,
    Minho Park$^{1}$,
    Junha Hyung$^{1}$,
    Jaegul Choo$^{1}$ \\
    $^{1}$KAIST AI, $^{2}$ Seoul National University\\
    {\small \texttt{\{keh0t0, kinamplify\}@kaist.ac.kr, } \texttt{kdh8156@snu.ac.kr}} \\
    {\small \texttt{\{m.park, sharpeeee, jchoo\}@kaist.ac.kr}}
}
\begin{document}

\twocolumn[{%
\renewcommand\twocolumn[1][]{#1}%
\maketitle
\begin{center}
\centering
\captionsetup{type=figure}
\includegraphics[width=1.0\textwidth]{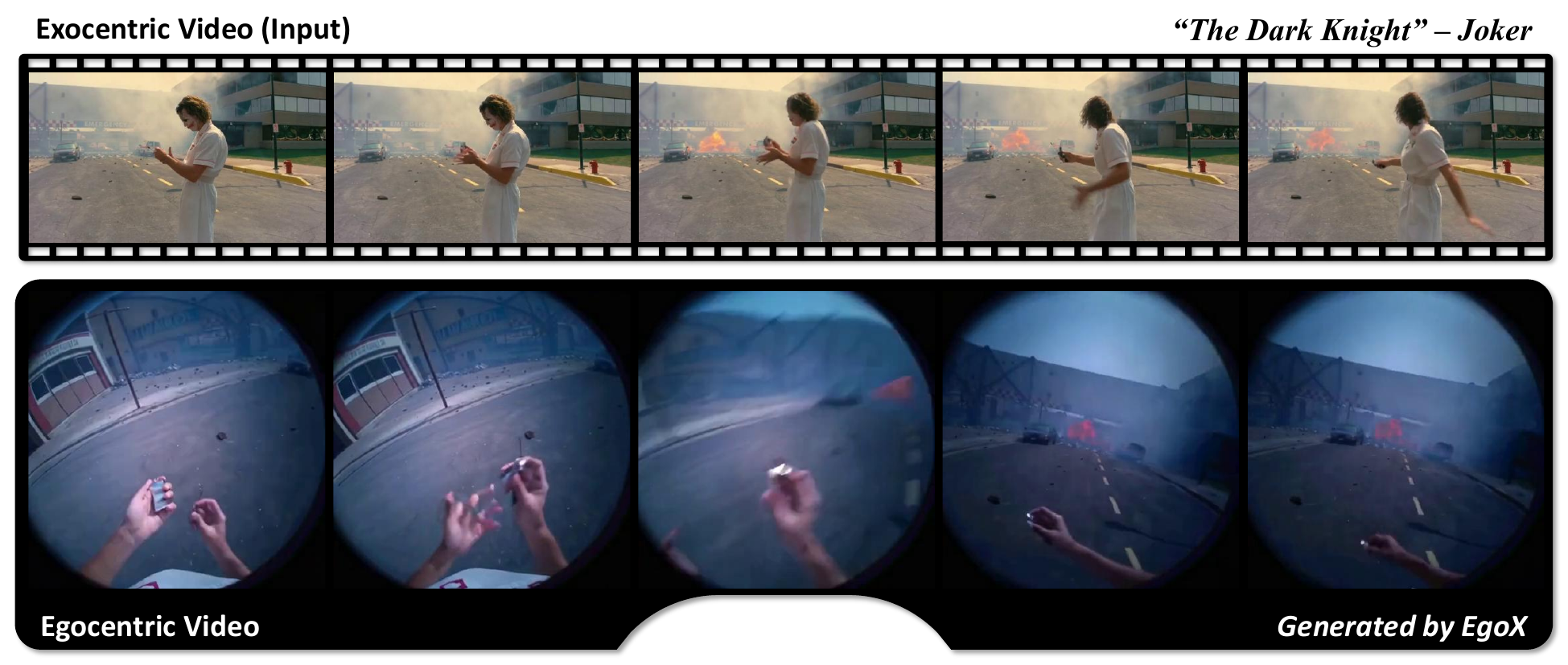}
\captionof{figure}{
Given a single exocentric video, \textbf{EgoX} generates what the scene would look like from the actor’s eyes. Shown with an in-the-wild clip from \textit{The Dark Knight}, our approach achieves realistic and generalizable egocentric generation.
}
\label{fig:1_teaser}
\end{center}
}]

\begin{abstract}
Egocentric perception enables humans to experience and understand the world directly from their own point of view. Translating exocentric (third-person) videos into egocentric (first-person) videos opens up new possibilities for immersive understanding but remains highly challenging due to extreme camera pose variations and minimal view overlap. This task requires faithfully preserving visible content while synthesizing unseen regions in a geometrically consistent manner. To achieve this, we present \textbf{EgoX}, a novel framework for generating egocentric videos from a single exocentric input.
EgoX leverages the pretrained spatio–temporal knowledge of large-scale video diffusion models through lightweight LoRA adaptation and introduces a unified conditioning strategy that combines exocentric and egocentric priors via width- and channel-wise concatenation.
Additionally, a geometry-guided self-attention mechanism selectively attends to spatially relevant regions, ensuring geometric coherence and high visual fidelity.
Our approach achieves coherent and realistic egocentric video generation while demonstrating strong scalability and robustness across unseen and in-the-wild videos.
\footnotetext{
\makebox[0.5cm][r]{}* indicates equal contributions.
}



\end{abstract}

\section{Introduction}
\label{sec:intro}

Don't you wish you could experience iconic scenes from films like \textit{The Dark Knight} as if you were the \textit{Joker} yourself? Exocentric-to-egocentric video generation makes this possible by converting a third-person scene into a realistic first-person perspective. This capability opens up new possibilities in the film industry, where viewers are no longer limited to passively watching a scene but can step into it and become the main character. They can become a superhero themselves or experience what it is like to play on the field as an MLB player.
Beyond entertainment, egocentric perspectives are crucial in fields such as robotics and AR/VR, where understanding how the world appears from the actor’s point of view enables better imitation, reasoning, and interaction~\cite{egocentricrobot, eyefaster}. This stems from the fact that humans perceive and interact with the world through a first-person, egocentric viewpoint. 

\begin{figure}[t]
  \centering
   \includegraphics[width=1.0\linewidth]{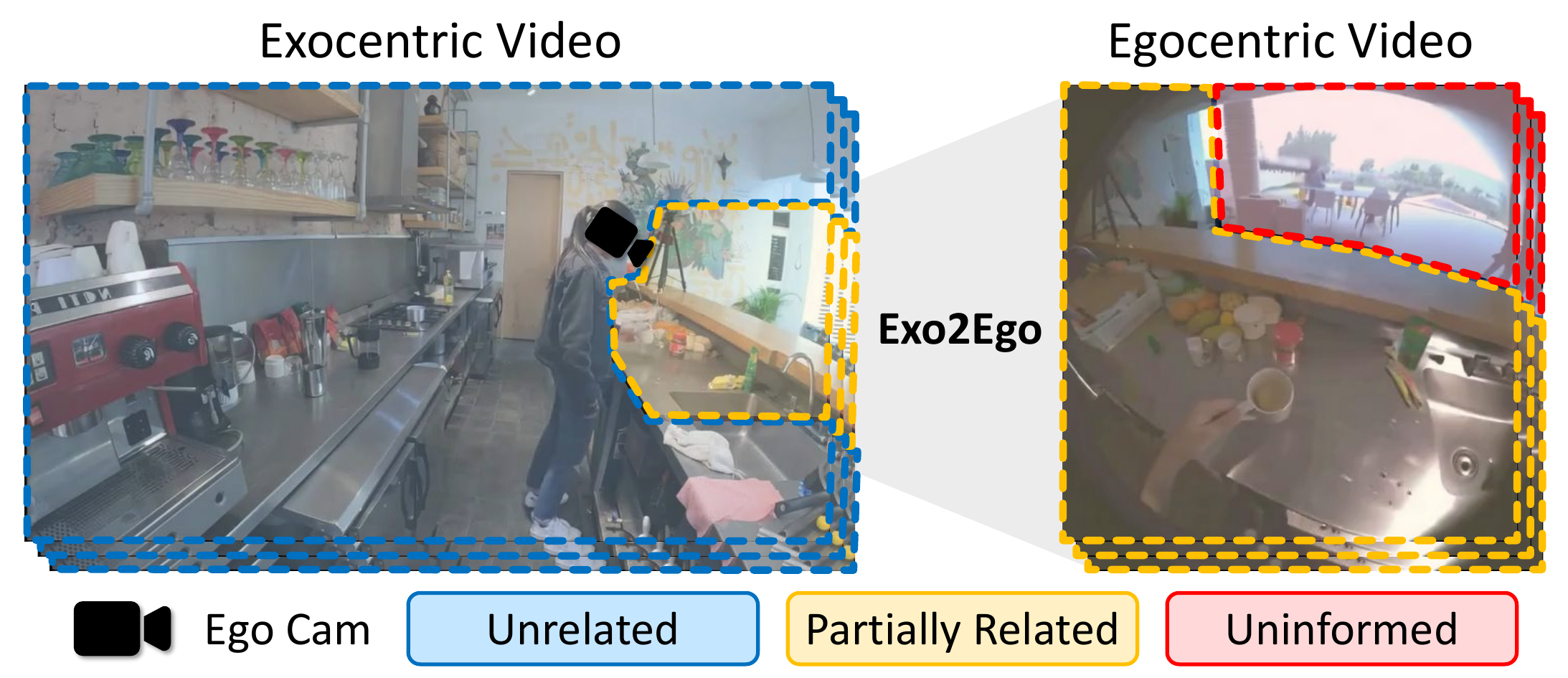}
    \caption{\textbf{Exo-to-Ego view generation example.} 
   The model has to preserve view-related content from the exocentric input, generate uninformed regions realistically, and ignore unrelated areas for consistent egocentric synthesis.}
   \label{fig:challenge}
   \vspace{-\baselineskip}
\end{figure}

However, generating such first-person perspectives is challenging, since the model must maintain scene consistency across views by reconstructing visible areas and realistically synthesizing unseen regions.
A straightforward way to achieve this is to use a camera control model.
Recent advances in camera control video generation models~\cite{trajectorycrafter, gen3c, voyager} have shown impressive performance in generating consistent views under moderate pose variations.
However, these methods primarily focus on modest viewpoint changes, whereas exocentric-to-egocentric video generation requires extreme camera pose translation that drastically alters the visible field of view.
This difference introduces two major challenges.
First, extreme viewpoint shifts result in large unseen regions that must be plausibly synthesized based on scene understanding rather than direct observation.
Second, only a small portion of the exocentric view corresponds to the egocentric perspective, making it crucial for the model to distinguish between view-related information that should be used as conditioning and unrelated content that should be suppressed.
As illustrated in~\cref{fig:challenge}, effective generation therefore requires selectively attending to meaningful regions while discarding irrelevant background areas and plausibly synthesizing uninformed regions in a geometrically consistent manner.
Therefore existing camera control models do not account for these challenges and thus often fail in exocentric-to-egocentric video generation.


Due to the inherent difficulty of this task, previous approaches often avoid generating the egocentric view from scratch or require additional inputs to simplify the problem.
EgoExo-Gen~\cite{egoexo-gen} takes both an exocentric video and the first egocentric frame as inputs to generate only the subsequent sequence.
Exo2Ego-V~\cite{exo2egov} utilizes four simultaneous exocentric camera views to capture richer spatial context and reduce the uninformed regions.

To address the limitations of previous approaches, we propose EgoX, a novel framework that generates egocentric video from a single exocentric video, achieving practical and generalizable egocentric generation from a single exocentric input.
Our method leverages the pretrained spatio–temporal knowledge of large-scale video diffusion models with minimal modification, enabling the model to plausibly synthesize unseen regions in a geometrically consistent manner.
Specifically, we design a unified conditioning strategy that combines exocentric views and egocentric priors through width-wise and channel-wise integration with clean latent representations, requiring only lightweight LoRA-based adaptation.
Furthermore, a geometry-guided self-attention allows the model to focus on spatially relevant regions while suppressing unrelated areas, leading to coherent and high-fidelity egocentric video generation.
By effectively leveraging pretrained weights, our approach produces high-quality egocentric videos and demonstrates strong generalization across diverse environments, including challenging in-the-wild scenarios, as illustrated in~\cref{fig:1_teaser}.

To summarize, the major contributions of our paper are as follows:

\begin{itemize}
    \item We propose a novel framework \textbf{EgoX} for synthesizing high-fidelity egocentric video from a \textit{single} exocentric video by effectively exploiting the pretrained video diffusion models.
    \item We design a unified conditioning strategy that jointly combines exocentric video and egocentric priors through width-wise and channel-wise integration, achieving robust geometric consistency and high-quality generation.
    \item We introduce a geometry-guided self-attention and clean latent representations that selectively focuses on view-relevant regions and enhances accurate reconstruction, leading to more coherent egocentric synthesis.
    \item Extensive qualitative and quantitative experiments demonstrate that \textbf{EgoX} outperforms previous approaches by a large margin, achieving \textit{state-of-the-art} performance on diverse and challenging exo-to-ego video generation benchmarks.
\end{itemize}

\section{Related Work}
\label{sec:related}
\subsection{Exo-to-Ego View Generation}
Prior works on exo-to-ego view generation have explored various conditioning mechanisms and task formulations to bridge the significant viewpoint gap.
Some approaches~\cite{exo2egov, exo2ego, egoworld} incorporate exocentric features by concatenating them channel-wise with the egocentric representation.
However, this method struggles with the fundamental lack of pixel-wise correspondence between the two viewpoints.
This spatial misalignment makes it difficult for the model to effectively leverage the conditioning information, often leading to a poor understanding of the scene geometry, which can result in overfitting or a degradation in output quality.
Other works, such as 4Diff~\cite{4diff}, employ cross-attention mechanisms to condition the generation on exocentric views.
This approach, however, prevents the utilization of powerful pretrained diffusion weights, limiting its generalizability and resulting in lower-quality synthesis.


To address these limitations, other methods utilize reference frames or multi-view conditions.
For instance, EgoExo-Gen~\cite{egoexo-gen} require the first egocentric frame to generate the rest of the sequence.
Exo2Ego-V~\cite{exo2egov} performs full video translation but relies on four exocentric video inputs and separately trained spatial and temporal modules, which limits its generalization and fails to fully exploit spatio–temporal priors.
In contrast, our model generalizes effectively using pretrained video diffusion weights while requiring only a single exocentric input.

\subsection{Video Diffusion Models}

Recent advancements in video diffusion models~\cite{wan, svd, sanavideo, cogvideox, ltx, cosmos} have led to significant improvements in generative quality, producing highly realistic and coherent video sequences. This has spurred a wide range of research exploring how to utilize these powerful generative capabilities in various applications~\cite{chen20254dnex, geo4d, spherediff, controlnet, vace}. A key area of this research focuses on conditional video generation, where the synthesis process is guided by specific inputs. Many works~\cite{vace, controlnet, standin, humo, tic-ft} have demonstrated successful control using conditions such as depth maps or static images.

Building on this, several methods have been proposed for camera-controlled video generation~\cite{camclonemaster, trajectorycrafter, recammaster}. These approaches can be broadly categorized into two main groups. The first group~\cite{recammaster, ac3d, worldmem, yume, yan} conditions the diffusion model directly on camera extrinsic parameters, often represented as raw matrices or Plücker coordinates. The second group~\cite{trajectorycrafter, gen3c, epic, realcami2v, voyager} first lifts the input video into an intermediate 3D representation, such as a point cloud. This 3D scene is then rendered from a new, user-specified camera pose, and the resulting image is used as a strong spatial condition to guide the final video generation.

However, existing methods for camera control are primarily designed for modest changes in viewpoint. They struggle to handle the extreme camera pose differences, a challenge that becomes particularly significant in exocentric-to-egocentric video generation. Our work addresses this critical gap by proposing a model capable of generating coherent egocentric videos from a significantly different exocentric perspective.

\begin{figure*}
  \centering
    \includegraphics[width=1.0\linewidth]{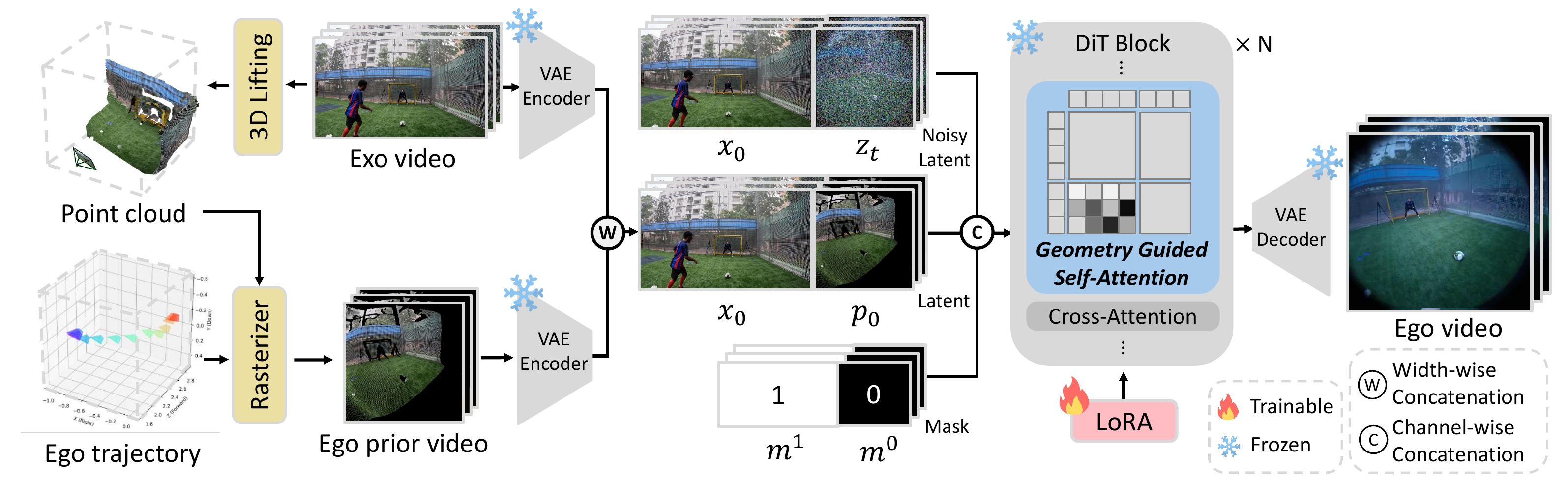}
  \caption{\textbf{Overall pipeline.} 
    Given an exocentric video input, we first lift it into a 3D point cloud and render the scene from the egocentric viewpoint to obtain the egocentric prior video.
    The clean exocentric video latent and the egocentric prior latent are combined via width-wise and channel-wise concatenation in the latent space, and then fed into a pretrained video diffusion model equipped with the proposed geometry-guided self-attention.}
  \label{fig:overview}
 \vspace{-\baselineskip}
\end{figure*}

\section{Method}
\label{sec:method}

Given an exocentric video sequence $X = \{X_i\}_{i=0}^{F}$ and egocentric camera pose $\phi= \{\phi_i\}_{i=0}^{F}$, the goal is to generate a corresponding egocentric video sequence $Y = \{Y_i\}_{i=0}^{F}$ that depicts the same scene from a first-person viewpoint.
The key challenge is to preserve the visible content in the exocentric view while synthesizing unseen regions in a geometrically consistent and realistic manner.
To this end, the exocentric sequence $X$ is first lifted into a 3D representation and rendered from the target egocentric viewpoint (\cref{subsec:3D_prior}), which becomes an egocentric prior video $P$.
Both $P$ and the original exocentric video $X$ are then provided as inputs to a video diffusion model (\cref{subsec:ic-lora}).
In addition, a geometry-guided self-attention (\cref{subsec:GGA}) is proposed to adaptively focus on view-consistent regions and enhance feature coherence across perspectives.

\subsection{Egocentric Point Cloud Rendering}
\label{subsec:3D_prior}

For this stage, we render an egocentric prior video $P \in \mathbb{R}^{F \times 3 \times H \times W}$ via point cloud rendering from the exocentric view. 
This prior provides both explicit pixel-wise RGB information and implicit camera trajectory cues that guide viewpoint alignment. 
Specifically, we first estimate a monocular depth map $D^m \in \mathbb{R}^{F \times H \times W}$ for each frame using a single-image depth estimator~\cite{moge2}, and a video-based depth map $D^v \in \mathbb{R}^{F \times H \times W}$ using a temporal depth estimator~\cite{vda}. 
Because $D^m$ is estimated independently per frame, depth values often exhibit slight inconsistencies across time. 
In contrast, $D^v$ produces a temporally smooth yet affine-invariant depth estimate. 
To combine the advantages of both, we temporally align $D^v$ with $D^m$. 
Following~\cite{vipe}, we optimize affine transformation parameters $\alpha, \beta$ using a momentum-based update strategy, yielding $\hat{\alpha} =\{\hat{\alpha}_f\}_{f=0}^{F}$ and $\hat{\beta}=\{\hat{\beta}_f\}_{f=0}^{F}$, which represent the per-frame affine transformations.
The final aligned depth is computed as:
\begin{equation}
D^f = \frac{1}{\hat{\alpha}/D^{v} + \hat{\beta}},
\end{equation}
where $D^f$ denotes the final aligned depth map. 
Dynamic objects are masked out so that only static background regions are used during both alignment and rendering. 
For further details, please refer to~\cite{vipe}.

After obtaining the aligned depth map $D^f$, we convert it into a 3D point cloud representation using the corresponding camera intrinsics. 
We then render the egocentric prior frames using a point cloud renderer~\cite{pytorch3d}:
\begin{equation}
P = \text{render}(X, D^f, \phi),
\end{equation}
where $X \in \mathbb{R}^{F \times 3 \times H \times W}$ is the exocentric RGB video and $\phi$ is egocentric camera poses. 



\subsection{Exo-to-Ego View Generation with VDM}
\label{subsec:ic-lora}



As illustrated in~\cref{fig:overview}, the model takes an exocentric video $X \in \mathbb{R}^{F \times 3 \times H \times W}$ and the egocentric prior video $P \in \mathbb{R}^{F \times 3 \times H \times W'}$ as conditioning inputs. 
Both inputs are encoded by a frozen VAE encoder, producing latent features $x_0 \in \mathbb{R}^{f \times c \times h \times w}$ and $p_0 \in \mathbb{R}^{f \times c \times h \times w'}$, respectively. 
These latents are then concatenated with the noisy latent $z_t \in \mathbb{R}^{f \times c \times h \times w'}$ to form the input of the diffusion model. 

The egocentric prior latent $p_0$ shares the same viewpoint as the target egocentric video and therefore preserves pixel-wise correspondence. 
We concatenate $p_0$ with $z_t$ along the channel dimension, providing viewpoint-aligned and temporally coherent guidance during generation. 
Although $p_0$ offers explicit geometric cues for the regions visible in the rendered ego view, it remains noisy and lacks substantial portions of the scene. 
To complement the missing information in the rendered egocentric view, we further use the exocentric video latent $x_0$ to provide broader scene context.
Since the viewpoint of $x_0$ differs from that of the noisy egocentric latent $z_t$, their features are not pixel-wise aligned. 
Therefore, we concatenate $x_0$ with $z_t$ along the width dimension, encouraging the model to infer cross-view correspondences and perform spatial warping implicitly.
Unlike~\cite{iclora}, which utilizes SDEdit~\cite{sdedit} by concatenating a noisy conditioning latent with a noisy target latent for conditional generation, our method concatenates the clean latent $x_0$ with the noisy $z_t$ throughout all denoising timesteps, while only $z_t$ is updated and $x_0$ remains fixed. 
This design encourages the model to consistently reference fine-grained details from $x_0$, enabling more accurate and reliable spatial warping.






The overall relation between inputs and outputs is defined as:
\begin{equation}
z_{t-1} = f_\theta({x_0, z_t} | {x_0, p_0} | {m^1, m^0}),
\end{equation}
where $f_\theta$ denotes a single-step denoising function of the VDM, $x_0$ is the exocentric video latent, $p_0$ is the egocentric prior latent, and $m$ is the binary mask specifying whether each spatial region is used for conditioning or for synthesis. Once the sampling is complete, we remove the exocentric part of the latent and decode only the egocentric part to obtain the final result.

\begin{figure*}[t]
  \centering
   \includegraphics[width=0.9\linewidth]{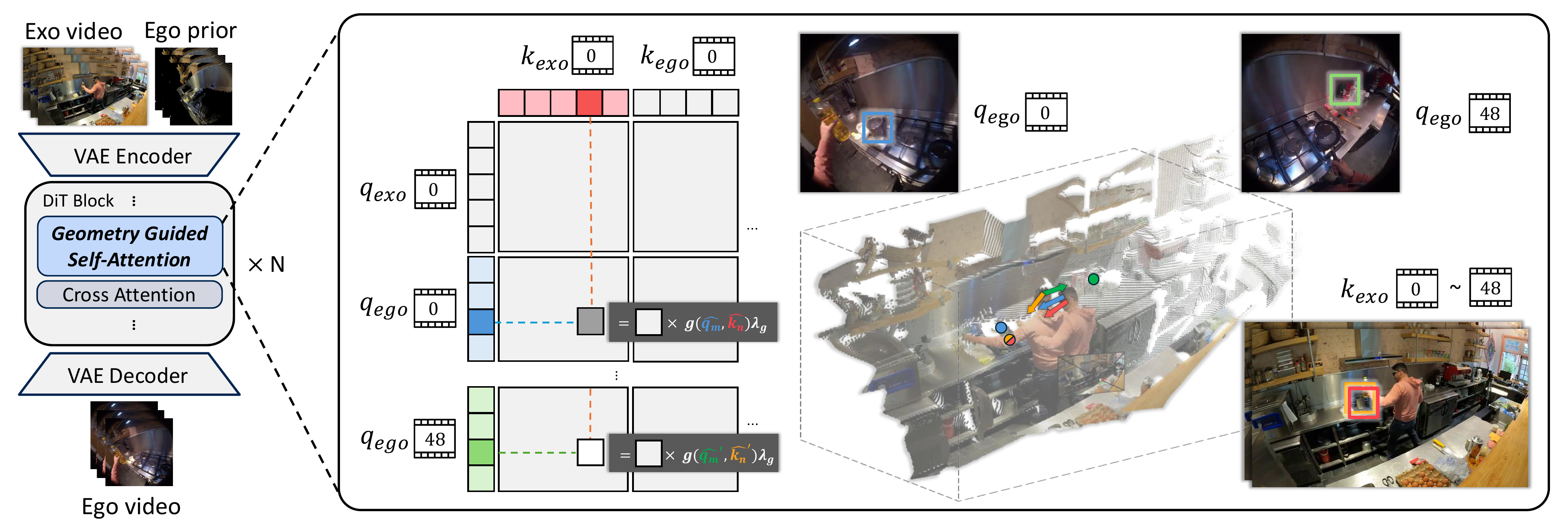}
   \caption{\textbf{Geometry-Guided Self-Attention Overview.} 
    3D direction similarities between egocentric queries and exocentric keys are used as an additive bias in the attention map, guiding the model to focus on geometrically aligned regions. Although the orange and red directions are the same key tokens, their directions differ due to different camera centers. The blue–red pairs have similar directions and thus receive higher scores, whereas the green–orange pairs have opposite directions and obtain lower scores.}
   \label{fig:gga}
 \vspace{-\baselineskip}
\end{figure*}

\subsection{Geometry-Guided Self-Attention}
\label{subsec:GGA}
As mentioned in~\cref{sec:intro}, the exocentric video condition includes irrelevant regions that can distract the model during exo-to-ego view generation. 
To address this, we introduce a Geometry-Guided Self-Attention (GGA) that adaptively emphasizes spatially corresponding regions between exocentric and egocentric representations. 
When egocentric query tokens $q_{\text{ego}} \in \mathbb{R}^{l\times c}$ attend to exocentric key tokens $k_{\text{exo}} \in \mathbb{R}^{l'\times c}$, the attention should jointly account for semantic similarity (i.e., appearance) and 3D spatial alignment. 
Ideally, tokens that are both semantically similar and geometrically aligned with the egocentric viewpoint should receive higher attention weights, while unrelated or misaligned regions are suppressed to ensure geometric consistency and realism in the generated views.

To achieve this, we leverage self-attention augmentation with 3D geometric cues. 
Using the 3D point cloud obtained in~\cref{subsec:3D_prior}, we compute 3D direction vectors from the ego camera centers $c = \{c_i\}_{i=0}^{F}, \, c_i \in \mathbb{R}^3$ in world space to each query and key token position, $\tilde{q}, \tilde{k} \in \mathbb{R}^3$. 
The unit direction vectors are defined as $ \hat{q} = \frac{\tilde{q} - c_i}{\|\tilde{q} - c_i\|_2}$, $\hat{k} = \frac{\tilde{k} - c_i}{\|\tilde{k} - c_i\|_2}, $. 
We then compute the cosine similarity between the two direction vectors and incorporate it into the attention computation as a multiplicative geometric prior.

Specifically, the modified attention logits are formulated as:
\begin{align}
s'_{m,n} &= s_{m,n} + \log\!\big(g(\hat{q}_m, \hat{k}_n) \cdot \lambda_{g}\big),        \\
g(\hat{a}, \hat{b}) &= \mathrm{cos\_sim}(\hat{a}, \hat{b}) + 1, \label{eq:geometry_bias_term}
\end{align}
where $s_{m,n} = \frac{q_m^\top k_n}{\sqrt{c}}$ denotes the standard attention logits~\cite{attention} and $\lambda_{g}$ is a hyperparameter that balances this geometry bias term defined in~\cref{eq:geometry_bias_term}. 
We add one to the cosine similarity term to ensure positive values before taking the logarithm. 

Finally, given an egocentric query $q_m$ and an exocentric key $k_n$, the attention weight $a_{m,n}$ is computed as:
\begin{align}
a_{m,n} &= \frac{
            \exp(s'_{m,n})
        }{
            \sum_{j=1}^{l} 
            \exp(s'_{m,j})
        } \\
        &=  \frac{
                \exp(s_{m,n})
                \, g(\hat{q}_m, \hat{k}_n) \, \lambda_{g}
            }{
                \sum_{j=1}^{l} 
                \exp(s_{m,j})
                \, g(\hat{q}_m, \hat{k}_j) \, \lambda_{g}
            }.
\end{align}
This formulation allows the attention mechanism to be explicitly guided by geometric alignment between query and key directions, improving spatial consistency and visual coherence across views.

In image generation, spatial relationships can be encoded by multiplying rotation matrices to each query and key before attention, as done in~\cite{rope, cape, prope, 4diff}. 
However, in video generation, the camera center of $q_{\text{ego}}$ changes at every frame, making it necessary to compute key directions relative to each query separately. 
This implies that the geometry bias term should be recomputed for every query–key pair within each frame’s attention operation. 
As illustrated in~\cref{fig:gga}, even $k_{\text{exo}}$ located at the same position (e.g. red) may have entirely different direction vectors (e.g. red and orange) depending on the camera pose.
To handle this, we compute all pairwise direction similarities between $k_{\text{exo}}$ and $q_{\text{ego}}$ and use this term as an additive bias attention mask, allowing us to reuse optimized attention kernels. 
This formulation provides a precise geometry-guided self-attention that effectively aligns exocentric and egocentric representations.

\begin{figure*}[t]
  \centering
   \includegraphics[width=1.0\textwidth]{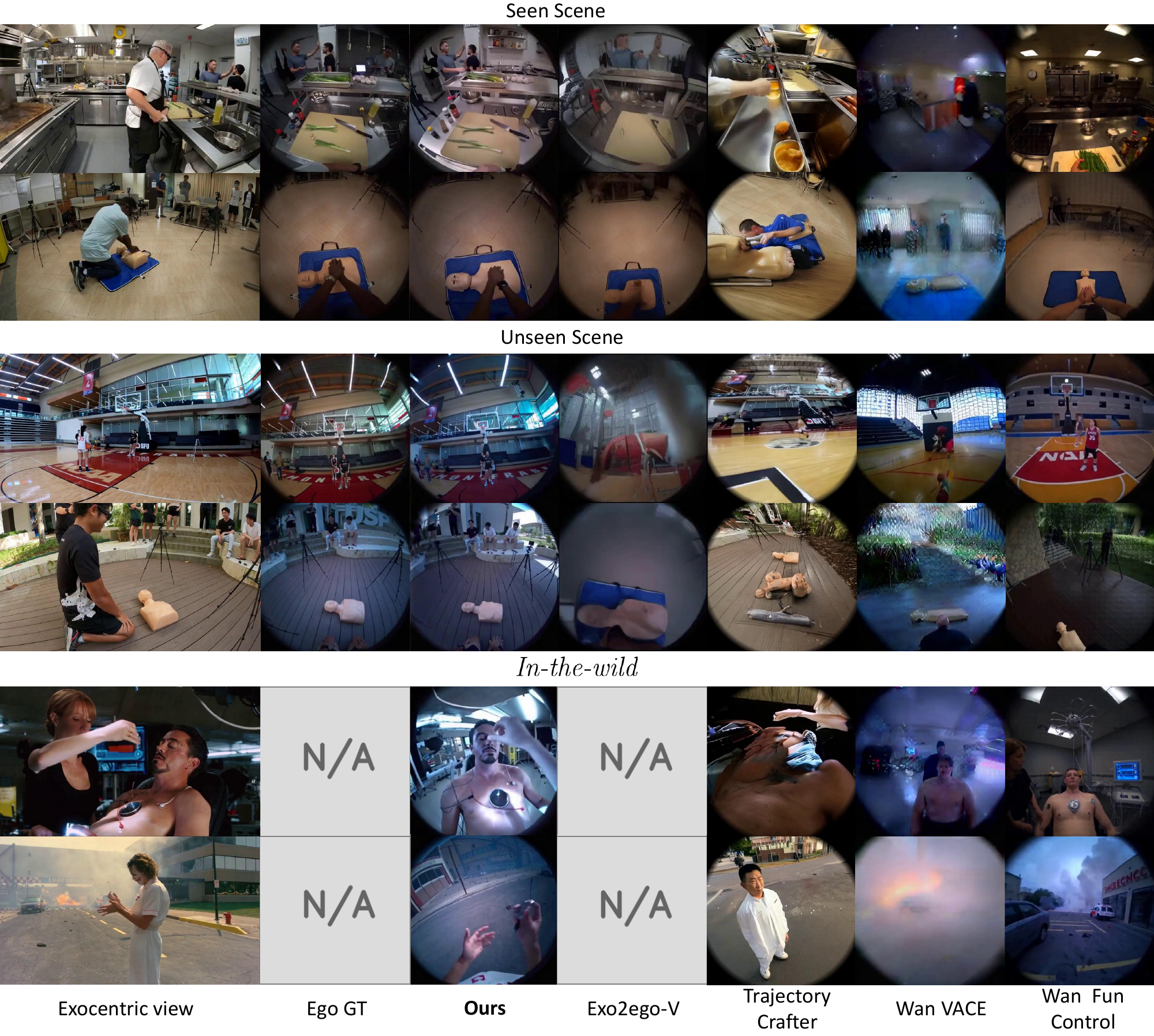}
   \caption{\textbf{Qualitative comparison.} Each example shows the exocentric input views and the corresponding generated egocentric views. While other methods fail to reconstruct realistic and coherent videos, our approach produces geometrically accurate and high-quality egocentric generations.
   N/A indicates that the result is unavailable either due to missing ground truth or the need for additional input views.}
   \label{fig:quali}
  \vspace{-\baselineskip}
\end{figure*}

\section{Experiments}
\label{sec:exp}

In the following sections, we aim to answer the following research questions that guide our experimental evaluation:
\begin{itemize}
    \item How does our method outperform existing baselines in both qualitative and quantitative evaluations? (\cref{subsec:quali}, \cref{subsec:quanti})
    \item How accurately does the model reconstruct regions visible in the exocentric view? (\cref{subsec:experimental_setup}, \cref{subsec:quanti})
    \item How well does the model generalize to unseen scenes and challenging in-the-wild videos? (\cref{subsec:quali}, \cref{subsec:quanti})
    \item How does each proposed component contribute to overall performance and generation quality? (\cref{subsec:ablation})

\end{itemize}

\begin{table*}[t]
\centering
\resizebox{0.97\textwidth}{!}{
\begin{tabular}{c|l|ccccccccccc}

\toprule
& & \multicolumn{4}{c}{Image Criteria} & \multicolumn{3}{c}{Object Criteria} & \multicolumn{3}{c}{Video Criteria} \\
\cmidrule(lr){3-6} \cmidrule(lr){7-9} \cmidrule(lr){10-13}
Scenarios & Method & PSNR $\uparrow$ & SSIM $\uparrow$ & LIPIS  $\downarrow$ & CLIP-I $\uparrow$ & \begin{tabular}[c]{@{}c@{}}Location \\ Error \end{tabular} $\downarrow$ & IoU $\uparrow$ & \begin{tabular}[c]{@{}c@{}}Contour\\ Accuracy\end{tabular} $\uparrow$ & FVD $\downarrow$ & \begin{tabular}[c]{@{}c@{}} Temporal \\ Flickering \end{tabular} $\uparrow$ & \begin{tabular}[c]{@{}c@{}}Motion \\ Smoothness \end{tabular} $\uparrow$  & \begin{tabular}[c]{@{}c@{}}Dynamic \\ Degree \end{tabular} $\uparrow$\\
\drule
\multirow{5}{*}{Seen Scenes}& Exo2Ego-V         & \underline{14.53} & 0.384             & \underline{0.569} & 0.774              & 156.66             & 0.074              & 0.364             & 622.47             & 0.960                 & 0.966          & \textbf{0.985}      \\
        & TrajectoryCrafter & 13.05             & 0.375             & 0.606             & 0.780              & \underline{100.74} & \underline{0.128}  & \underline{0.427} & 546.09            & 0.960              & 0.980             & 0.947    \\
        & Wan Fun Control   & 12.25             & \underline{0.463} & 0.617             & 0.810              & 112.57             & 0.076              & 0.417             & 595.07            & 0.968              & 0.980             & 0.901     \\
        & Wan VACE          & 12.95             & 0.413             & 0.626             & \underline{0.829}  & 109.62             & 0.114              & 0.376             & 508.69            & \textbf{0.989}     & \textbf{0.994}    & 0.673     \\
        & EgoX (Ours)       & \textbf{16.05}    & \textbf{0.556}    & \textbf{0.498}    & \textbf{0.896}     & \textbf{61.81}     & \textbf{0.363}     & \textbf{0.546}    & \textbf{184.47}   & \underline{0.977}  & \underline{0.990} & \underline{0.974}\\
\drule
\multirow{5}{*}{Unseen Scenes}& Exo2Ego-V         & 12.70             & \underline{0.439} & \underline{0.597} & 0.679              & 214.32            & 0.003             & 0.296            & 1283.50             & 0.971         & 0.976          & \underline{0.978} \\
        & TrajectoryCrafter & 12.24             & 0.297             & 0.619             & 0.778              & 192.16            & 0.039             & 0.301            & \underline{821.71}& 0.966             & 0.984             & 0.944  \\
        & Wan Fun Control   & \underline{13.59} & 0.439             & 0.604             & 0.799              & 191.40            & 0.042             & 0.329            & 968.78            & 0.971             & 0.985             & 0.944  \\
        & Wan VACE          & 12.17             & 0.345             & 0.638             & \underline{0.820}  & 191.97            & 0.038             & 0.314            & 1045.45           & \textbf{0.995}    & \textbf{0.996}    & 0.427  \\
        & EgoX (Ours)       & \textbf{14.38}    & \textbf{0.457}    & \textbf{0.552}    & \textbf{0.877}     & \textbf{149.93}   & \textbf{0.092}    & \textbf{0.481}   & \textbf{440.64}   & \underline{0.981} & \underline{0.992} & \textbf{0.989}  \\
\bottomrule

\end{tabular}
}
\caption{
\textbf{Quantitative Results.} 
Comparison on image, object, and video metrics. Our method achieves the best overall performance, with Wan VACE showing higher video scores due to static outputs. \textbf{Best} results are highlighted in bold, and \underline{second-best} results are underlined.
}
\label{tab:exp_quanti}
\end{table*}

\subsection{Experimental Setup}
\label{subsec:experimental_setup}
\paragraph{Implementation Details.}
To support channel-wise concatenation of noisy latent and ego prior latent, we adopt the inpainting variant of Wan 2.1 (14B) Image-to-Video model~\cite{wan} as our base model. 
We fine-tuned the model using LoRA (rank = 256) with a batch size of 1, and a single day on 8 H200 (140 GB) GPUs.
For the dataset, we curated 4,000 clips from Ego-Exo4D~\cite{ego4d} covering diverse scenes and actions, using 3,600 clips for training and 400 for testing.
Additionally, we collected 100 unseen clips that are not included in the training set to evaluate generalization performance.
More detailed information can be found in ~\cref{Appen:detail}.



\begin{table*}[t]
\centering
\resizebox{0.97\textwidth}{!}{
\begin{tabular}{l|ccccccccccccc}

\toprule
& \multicolumn{4}{c}{Image Criteria} & \multicolumn{3}{c}{Object Criteria} & \multicolumn{3}{c}{Video Criteria} \\
\cmidrule(lr){2-5} \cmidrule(lr){6-8} \cmidrule(lr){9-12}
Method & PSNR $\uparrow$ & SSIM $\uparrow$ & LIPIS  $\downarrow$ & CLIP-I $\uparrow$ & \begin{tabular}[c]{@{}c@{}}Location \\ Error \end{tabular} $\downarrow$ & IoU $\uparrow$ & \begin{tabular}[c]{@{}c@{}}Contour\\ Accuracy\end{tabular} $\uparrow$ & FVD $\downarrow$ & \begin{tabular}[c]{@{}c@{}} Temporal \\ Flickering \end{tabular} $\uparrow$ & \begin{tabular}[c]{@{}c@{}}Motion \\ Smoothness \end{tabular} $\uparrow$  & \begin{tabular}[c]{@{}c@{}}Dynamic \\ Degree \end{tabular} $\uparrow$\\
\drule
EgoX (Ours)               & \textbf{16.05}    & \textbf{0.556}    & \textbf{0.498}    & \underline{0.896} & \textbf{61.81}     & \textbf{0.363}    & \textbf{0.546}    & \textbf{184.47}& \textbf{0.977}    & \underline{0.989 }& \textbf{0.974}   \\
w/o GGA                   & 14.77             & \underline{0.539} & \underline{0.530} & \textbf{0.897}    & \underline{64.30}  & \underline{0.326} & \underline{0.538} & 254.08         & 0.969             & 0.987             & \underline{0.877}\\
w/o Ego prior             & 13.67             & 0.479             & 0.573             & 0.864             & 90.70              & 0.417             & 0.464             & 211.50         & \underline{0.974} & \textbf{0.990}    & 0.802\\
w/o clean latent          & \underline{15.07} & 0.528             & 0.540             & 0.861             & 70.17              & 0.376             & 0.506             & 343.33         & 0.963             & 0.986             & 0.864\\

\bottomrule

\end{tabular}
}
\caption{
\textbf{Ablation Study Results.} Performance comparison by removing each core component of our framework. 
The full model achieves the best results, while excluding geometry-guided self-attention, ego prior, or clean latent conditioning causes performance degradation. \textbf{Best} results are highlighted in bold, and \underline{second-best} results are underlined.
}
\label{tab:exp_ablation}
\end{table*}

\begin{figure*}[t]
  \centering
   \includegraphics[width=0.9\textwidth]{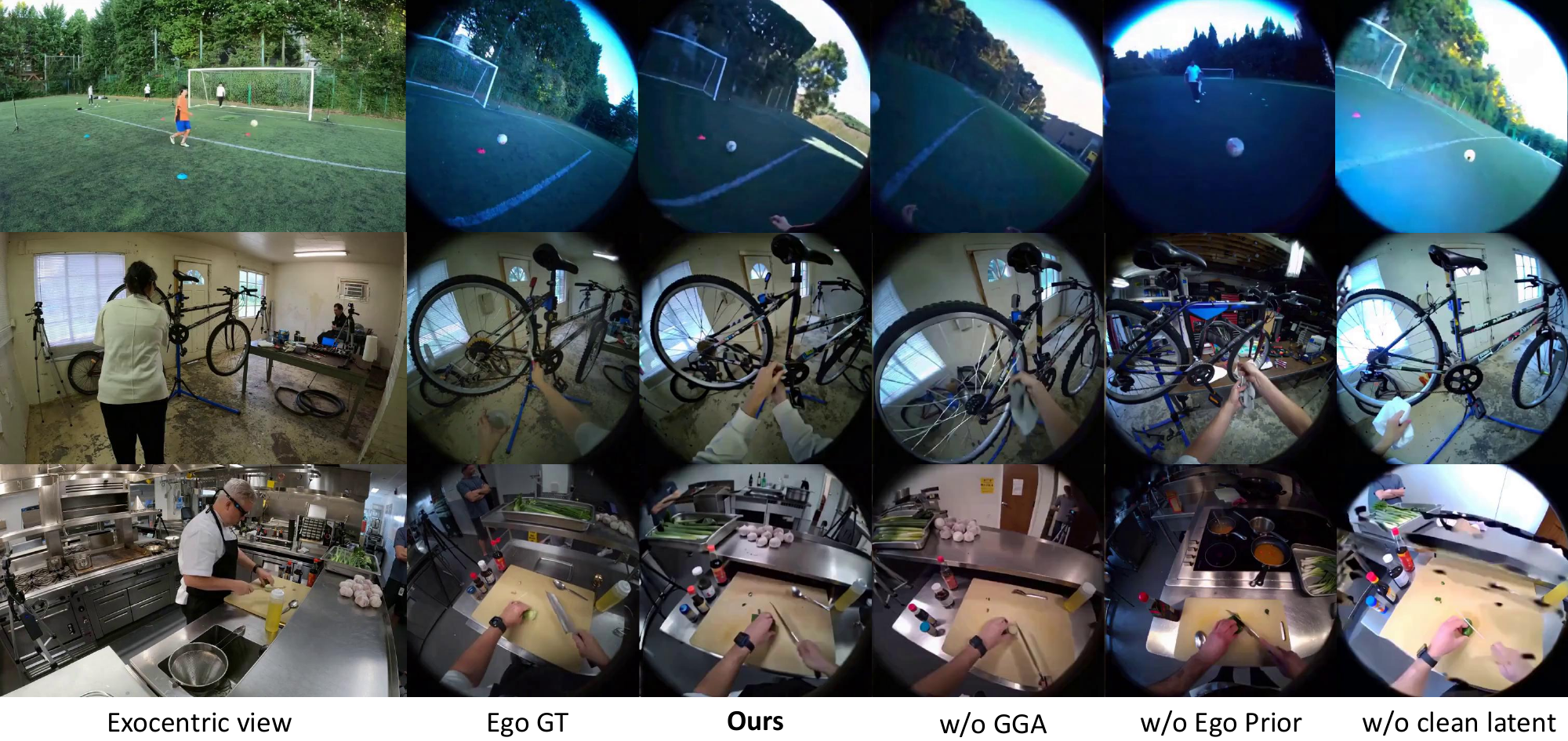}
    
   \caption{\textbf{Ablation qualitative comparison.} Visual results when removing each core component. Removing any single component, GGA, the egocentric prior, or the clean latent representation, results in degraded generation quality and geometric consistency.}
   \label{fig:ablation}

\end{figure*}

\paragraph{Baselines.}

Among existing exocentric-to-egocentric video generation approaches, Exo2Ego-V~\cite{exo2egov} and EgoExo-Gen~\cite{egoexo-gen} serve as representative baselines.
We adopt Exo2Ego-V as our primary baseline, as EgoExo-Gen does not provide publicly available implementation.
With the rapid progress in conditional video generation and camera control models, several recent methods have demonstrated performance comparable to or even surpassing Exo2Ego-V.
Therefore, we additionally included Trajectory Crafter~\cite{trajectorycrafter}, a state-of-the-art camera control model, as well as Wan Fun Control~\cite{VideoX_Fun} and Wan VACE~\cite{vace}, which offer distinct conditioning approach.
Wan Fun Control applies channel-wise concatenation for conditioning, and Wan VACE employs an auxiliary conditioning network, providing diverse points of comparison for our method. For the fair comparison, we finetuned these baselines using the same training dataset as ours.

\paragraph{Evaluation Metrics.}
To evaluate the quality of generated videos, we employed three types of criteria.

\begin{itemize}[leftmargin=1.2em]
    \item \textbf{Image Criteria.}
    We measured PSNR, SSIM, LPIPS, and CLIP-I to assess how closely each generated frame matches the ground-truth distribution.

    \item \textbf{Object Criteria.}
    Following the object-level evaluation protocol of Ego-Exo4D~\cite{ego-exo4d}, we assessed object-level consistency between the generated egocentric video and the ground truth. We used SAM2~\cite{sam2} to segment and track objects and DINOv3~\cite{dinov3} to establish correspondences. For each matched object, we evaluated center-location error, Intersection-Over-Union(IoU), and Contour Accuracy to measure spatial alignment and boundary fidelity.
    
    \item \textbf{Video Criteria.}
    We measured FVD~\cite{fvd} to evaluate how closely the generated video aligns with the ground-truth distribution. In addition, we assessed VBench~\cite{vbench}-Temporal Flickering, Motion Smoothness, and Dynamic Degree to quantify temporal stability and motion quality.
\end{itemize}

\subsection{Qualitative Results}
\label{subsec:quali}
\cref{fig:quali} visualizes the qualitative comparisons between our method and the baselines. Note that in the \textit{in-the-wild} scenario, ground-truth egocentric videos are unavailable, and Exo2Ego-V is also not applicable since only a single exocentric video is provided, which does not meet its four-view input requirement.
Exo2Ego-V fails to generate high-fidelity frames even when using four exocentric inputs, whereas our model achieves superior visual quality and generalizes well to unseen scenes from only a single exocentric view.
Trajectory Crafter struggles with large camera translations, producing spatial distortions and temporal inconsistencies.
Both Wan VACE and Wan Fun Control fail to effectively utilize the exocentric conditioning input, resulting in mismatched geometry, degraded realism, and the inclusion of irrelevant exocentric content in the egocentric view.
Overall, these results demonstrate that our model effectively leverages pretrained video diffusion knowledge to generate geometrically accurate, visually coherent, and highly realistic egocentric videos, maintaining strong performance even under challenging in-the-wild conditions.
More qualitative results, including temporally aligned visualizations, can be found in ~\cref{Appen:results}.



\begin{figure}[t]
  \centering
   \includegraphics[width=0.9\linewidth]{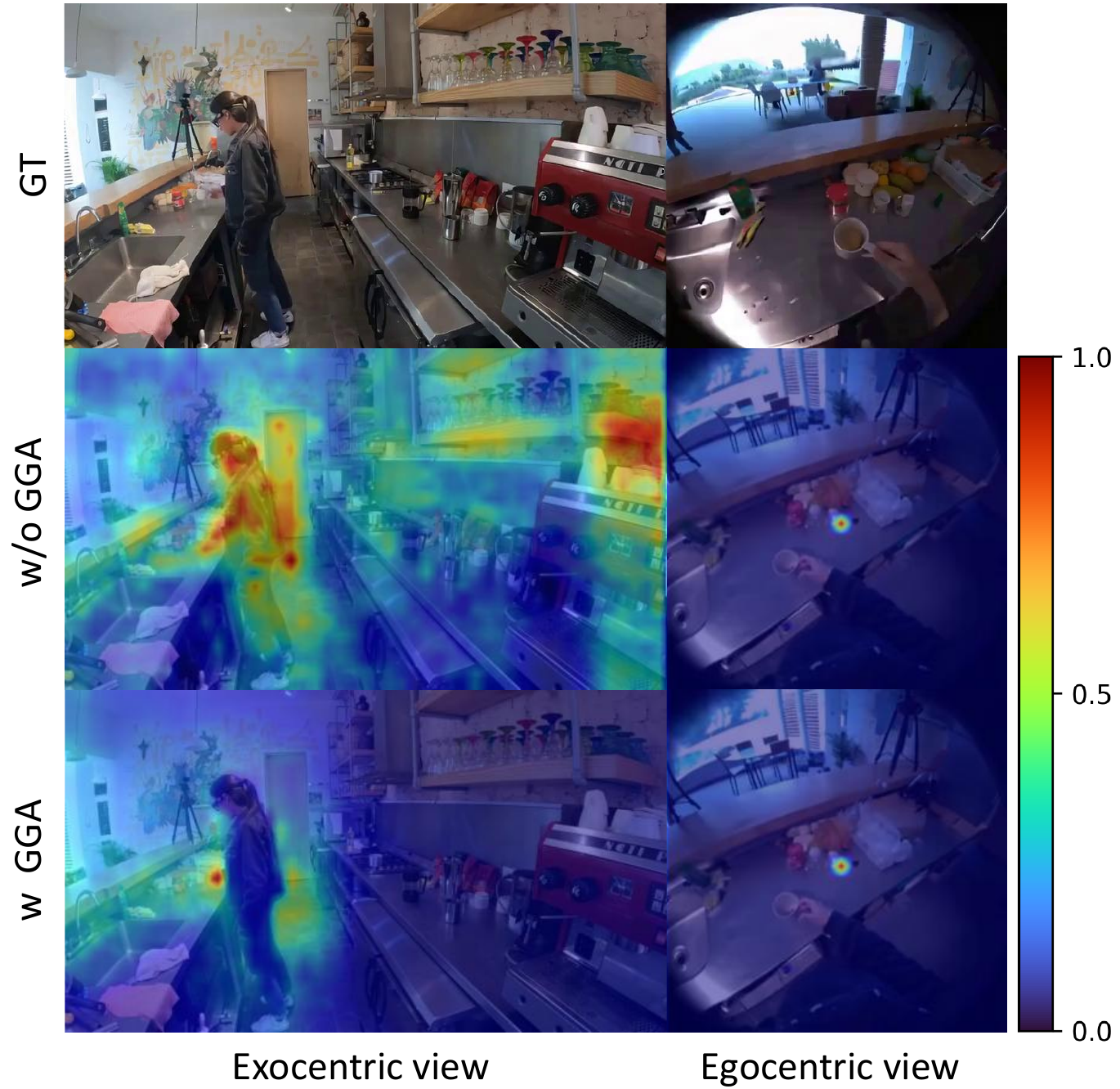}
    
   \caption{\textbf{Attention map visualization.}
   Visualization of the attention weights when querying the center token of the egocentric view. Without GGA, the model attends to unrelated regions, whereas with GGA, attention is concentrated on related regions, highlighting improved spatial alignment.
   }
   \label{fig:attention_map}
  \vspace{-\baselineskip}
\end{figure}

\subsection{Quantitative Results}
\label{subsec:quanti}

As shown in~\cref{tab:exp_quanti}, our method achieves the best overall performance across both image and object criteria.
In particular, we observe a significant performance gap in the object-based criteria, indicating that our approach preserves scene geometry and object consistency more effectively than other baselines.
While image-level scores may appear slightly lower due to the inherent challenge of synthesizing unseen regions that differ from the ground-truth egocentric view, our method still achieves the best results across all image metrics.
For video-based metrics, Wan VACE records the highest temporal smoothness and flicker scores.
However, this is largely attributed to its generation of overly static videos with limited motion, resulting in low dynamic degree.
In contrast, our model produces temporally coherent and visually dynamic sequences, demonstrating a better balance between spatial fidelity and motion realism.

\subsection{Ablation Study}
\label{subsec:ablation}


We conducted ablation studies to evaluate the contribution of each core component in our framework, including the geometry-guided self-attention (GGA), the egocentric prior conditioning, and the clean latent representation.
For each ablation variant, one component was removed while keeping all other settings identical. Quantitative evaluations were performed on the seen scene subset to ensure a controlled comparison.
As shown in~\cref{fig:ablation} and~\cref{tab:exp_ablation}, removing any of these components results in a noticeable performance drop, both qualitatively and quantitatively. 
Without GGA, the model fails to maintain geometric alignment, attending to broad and unrelated regions, which leads to spatial inconsistency.
Without the egocentric prior, the model lacks explicit pixel-wise and camera trajectory information, thus struggling to follow the correct viewpoint and producing visually implausible frames.
Without the clean latent, the exocentric latent is concatenated in a noisy state, which blurs fine-grained details. As a result, the target latent fails to preserve these details, leading to missing or degraded object structures. In the last row of~\cref{fig:ablation}, for instance, the model does not generate the spoon or the small circular ingredients on the cutting board that appear in the ground-truth egocentric view. 

To further demonstrate the effectiveness of the geometry-guided self-attention, we visualize the attention maps queried by egocentric tokens.
As shown in~\cref{fig:attention_map}, without GGA, the model attends to broad irrelevant regions, while with GGA, it sharply focuses on view-relevant areas, reinforcing geometric coherence and stabilizing feature alignment.
Additional ablation studies are provided in~\cref{Appen:ablation}.

\section{Conclusion}
\label{sec:con}
We introduce \textbf{EgoX}, the first framework capable of generating egocentric videos from a single exocentric input while achieving strong generalization across diverse scenes. 
Our method introduces a unified conditioning strategy that combines exocentric and egocentric priors via width- and channel-wise concatenation for effective global context and viewpoint alignment, while leveraging lightweight LoRA-based adaptation to preserve the pretrained video diffusion model’s spatio-temporal reasoning ability.
Furthermore, clean latent representations and geometry-guided self-attention enable the model to selectively focus on spatially relevant regions and maintain geometric consistency, resulting in coherent and high-fidelity egocentric generation. Despite its effectiveness, our current framework requires an egocentric camera pose as input. Although this information can be provided interactively by users, incorporating an automatic head-pose estimation module would be a valuable future direction.




{
    \small
    \bibliographystyle{ieeenat_fullname}
    \bibliography{main}
}

\clearpage
\setcounter{page}{1}
\maketitlesupplementary

\renewcommand{\thesection}{\Alph{section}}

\section{Implementation Detail}
\label{Appen:detail}

\begin{table*}[t]
\centering
\resizebox{0.97\textwidth}{!}{
\begin{tabular}{l|ccccccccccccc}

\toprule
& \multicolumn{4}{c}{Image Criteria} & \multicolumn{3}{c}{Object Criteria} & \multicolumn{3}{c}{Video Criteria} \\
\cmidrule(lr){2-5} \cmidrule(lr){6-8} \cmidrule(lr){9-12}
Method & PSNR $\uparrow$ & SSIM $\uparrow$ & LIPIS  $\downarrow$ & CLIP-I $\uparrow$ & \begin{tabular}[c]{@{}c@{}}Location \\ Error \end{tabular} $\downarrow$ & IoU $\uparrow$ & \begin{tabular}[c]{@{}c@{}}Contour\\ Accuracy\end{tabular} $\uparrow$ & FVD $\downarrow$ & \begin{tabular}[c]{@{}c@{}} Temporal \\ Flickering \end{tabular} $\uparrow$ & \begin{tabular}[c]{@{}c@{}}Motion \\ Smoothness \end{tabular} $\uparrow$  & \begin{tabular}[c]{@{}c@{}}Dynamic \\ Degree \end{tabular} $\uparrow$\\
\drule
EgoX (Ours)     & \textbf{14.38}    & \textbf{0.457}    & \textbf{0.552}    & 0.877             & \textbf{149.93}     & \textbf{0.092}    & \textbf{0.481}    & \textbf{440.64}   & \textbf{0.9813} & \textbf{0.9923}     & \textbf{0.989}  \\
w/o GGA         & 13.27             & 0.432             & 0.587             & \textbf{0.880}    & 154.27              & 0.089             & 0.400             & 522.67            & 0.9812          & 0.9921               & 0.955            \\
w/o Ego prior   & 13.01             & 0.401             & 0.581             & 0.855             & 171.95              & 0.059             & 0.351             & 523.00            & 0.9742           & 0.9908             & 0.843       \\
w/o Clean latent& 14.06             & 0.426             & 0.571             & 0.828             & 169.20              & 0.063             & 0.328             & 695.01            & 0.9811           & 0.9917             & 0.876        \\

\bottomrule

\end{tabular}
}
\caption{
\textbf{Ablation Study Results on Unseen Scenes.}
The performance trends on unseen scenes are consistent with those observed on seen scenes.
\textbf{Best} results are highlighted in bold.
}
\label{tab:Appen_exp_ablation}
\end{table*}

\subsection{GGA Implementation Detail}
Applying Geometry-Guided Self-Attention (GGA) directly in pixel space is not feasible because the diffusion model operates in the latent space.
Therefore, we compute 3D direction vectors at the pixel level and downsample them by averaging over each 4$\times$16$\times$16 patch, matching the VAE downsampling factor include temporal dimension. The resulting patch-level direction vectors are used as geometric cues in the latent-space attention.

These geometric terms are precomputed once before the model inference to avoid runtime overhead.
Additionally, applying the geometry-guided bias to all attention layers simultaneously would significantly increase memory usage and computational cost.
To address this, we separately apply attention kernels for ego-to-exo and exo-to-ego attention, enabling efficient integration of geometric bias without exceeding memory constraints.

\begin{figure}[t]
  \centering
   \includegraphics[width=1.0\linewidth]{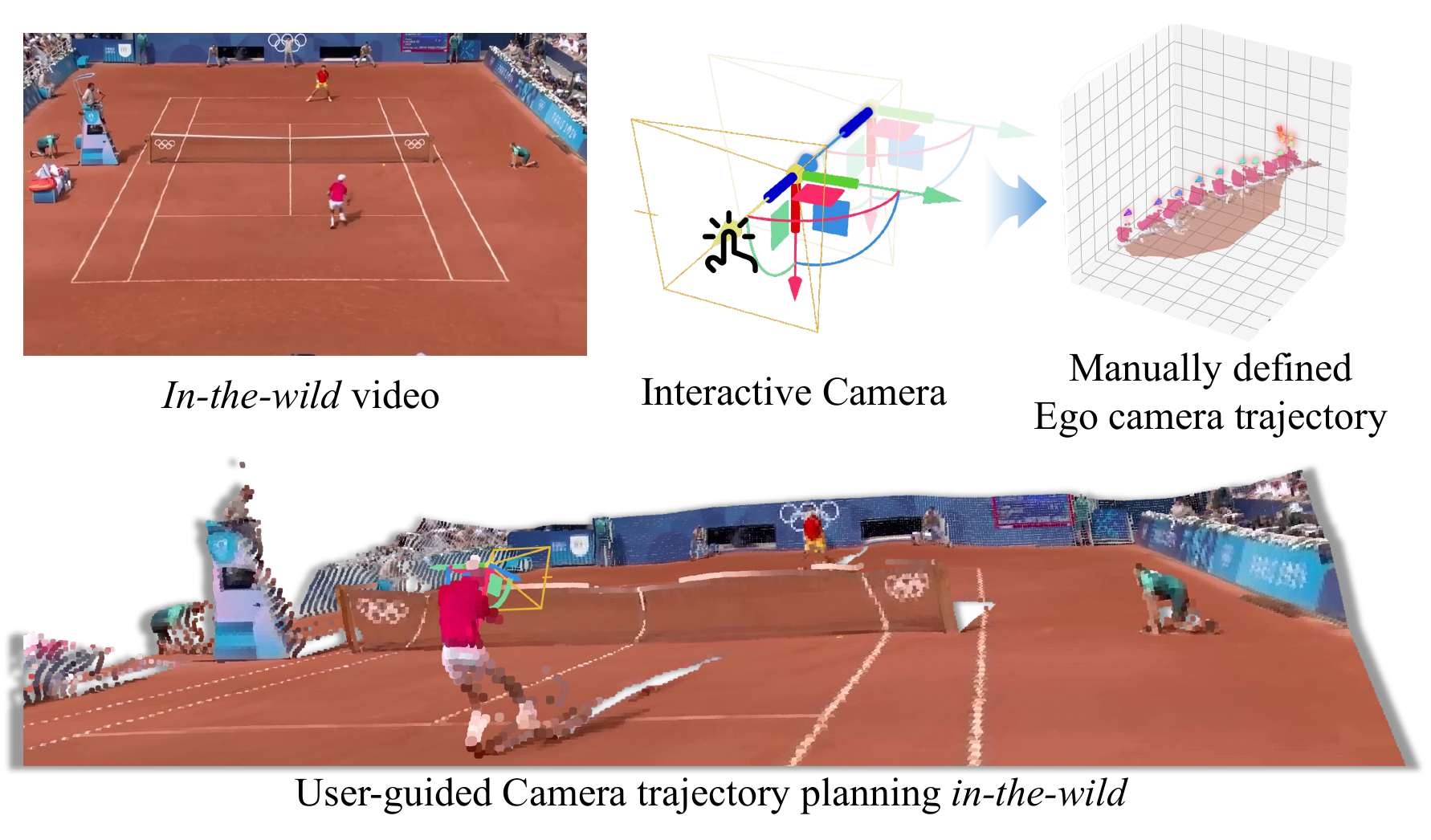}
    
   \caption{\textbf{In-the-wild Ego camera}. The ego camera for the in-the-wild example was obtained by interactively determining its extrinsic parameters using Viser~\cite{viser}.}
   \label{fig:A_ego_traj}
\end{figure}

\begin{figure}[t]
  \centering
   \includegraphics[width=1.0\linewidth]{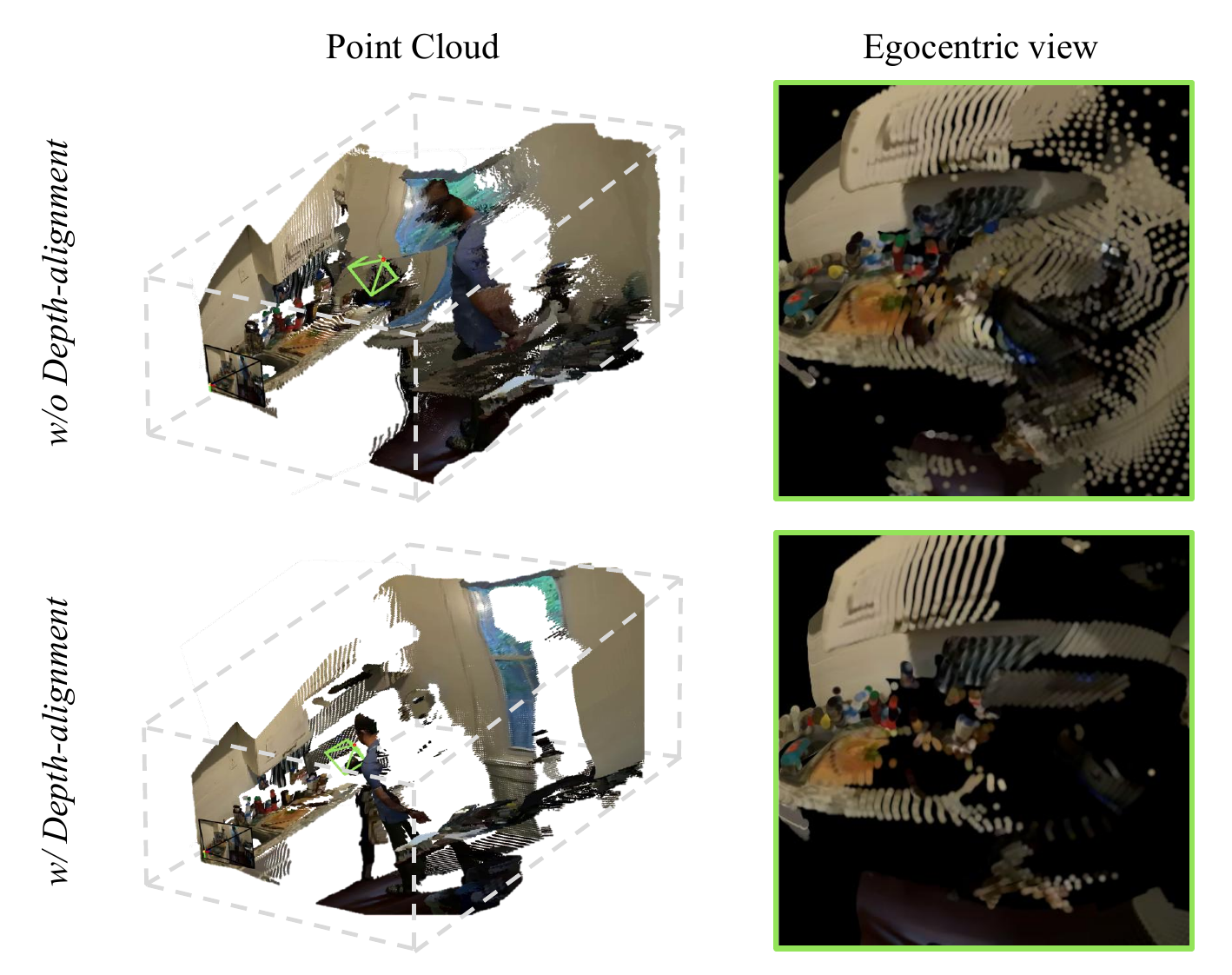}
    
   \caption{\textbf{Depth align comparison.} The above egocentric view is rendered from 3D point clouds across all frames. Without depth alignment, the inconsistent depth values between frames lead to unstable and unexpected camera movements.}
   \label{fig:A_vda}
\end{figure}

\subsection{Ego Camera Pose for In-the-wild Example}
Unlike the EgoExo4D~\cite{ego-exo4d} dataset, where ground-truth egocentric camera poses are provided, our in-the-wild examples do not include any ego camera pose annotations.
To obtain the required egocentric poses for rendering, we manually determined the camera extrinsics using the 3D visualization toolkit Viser~\cite{viser}.
Specifically, we lifted the exocentric video into a 3D point cloud and interactively selected the camera pose that best matches the expected egocentric viewpoint, as illustrated in~\cref{fig:A_ego_traj}.
As mentioned in~\cref{sec:con}, incorporating an automatic head-pose estimation module would be a valuable future extension.
Potential options include video-based head-pose trackers~\cite{HeadPoseEstimator} or SMPL~\cite{smpl}-based pose estimators, which could eliminate manual intervention and enable fully automatic exocentric-to-egocentric generation.

\subsection{Evaluation Detail}
In this section, we detail the evaluation procedure used to compute the Object Criteria, leveraging SAM2~\cite{sam2} for object segmentation and DINOv3~\cite{dinov3} for appearance-based object matching.

\subsubsection*{Object Criteria}
For each video, we perform object segmentation using SAM2 to obtain all valid object regions. For every detected object, we extract its bounding box and corresponding contour mask. Each object region is then cropped according to its bounding box and encoded into a feature vector $\mathbf{f} \in \mathbb{R}^d$ using a pretrained DINOv3 model.

To establish correspondences between the ground-truth egocentric video and the generated output, we compute cosine similarities for all possible pairs of object embeddings:
\begin{equation}
    s_{i,j} = 
    \frac{
        \mathbf{f}_i^{\text{GT}} \cdot \mathbf{f}_j^{\text{model}}
    }{
        \left\lVert \mathbf{f}_i^{\text{GT}} \right\rVert_2
        \left\lVert \mathbf{f}_j^{\text{model}} \right\rVert_2
    }.
\end{equation}
A pair $(i, j)$ is considered a valid correspondence only if it satisfies a high-confidence appearance threshold $s_{i,j} \ge \tau_{\text{sim}}$, where we set $\tau_{\text{sim}} = 0.9$.  
These high-confidence matched object pairs form the basis for all downstream object-level metrics.

\paragraph{Location Error.}
For a valid matched pair, spatial alignment is measured using the Euclidean distance between the centers of the two bounding boxes. Let $\mathbf{c}_i^{\text{GT}}$ and $\mathbf{c}_j^{\text{model}}$ denote their centers. The location error is computed as:
\begin{equation}
    \mathcal{E}_{i,j}^{\text{loc}}
    = 
    \left\lVert 
        \mathbf{c}_i^{\text{GT}} - \mathbf{c}_j^{\text{model}}
    \right\rVert_2.
\end{equation}
Lower values indicate better spatial consistency.

\paragraph{Bounding Box IoU.}
To measure coarse geometric consistency, we compute the Intersection over Union (IoU) between the two bounding boxes:
\begin{equation}
    \mathrm{IoU}_{i,j}
    =
    \frac{
        \mathrm{Area}(B_i^{\text{GT}} \cap B_j^{\text{model}})
    }{
        \mathrm{Area}(B_i^{\text{GT}} \cup B_j^{\text{model}})
    }.
\end{equation}
Higher IoU indicates closer agreement in object position and scale.

\paragraph{Contour Accuracy.}
To evaluate fine-grained geometric consistency, we measure contour-level similarity using the object contours extracted by SAM2. For each matched object pair, SAM2 produces a contour mask, which we denote as $C_i^{\text{GT}}$ and $C_j^{\text{model}}$ for the ground-truth and generated frames, respectively.  
The contour IoU is then computed as:
\begin{equation}
    \mathrm{IoU}^{\text{contour}}_{i,j}
    =
    \frac{
        \left| C_i^{\text{GT}} \cap C_j^{\text{model}} \right|
    }{
        \left| C_i^{\text{GT}} \cup C_j^{\text{model}} \right|
    }.
\end{equation}
This metric captures whether the object shape is preserved beyond the coarse bounding-box alignment.


\subsection{Text Prompts}
\begin{figure*}[t]
  \centering
   \includegraphics[width=1.0\linewidth]{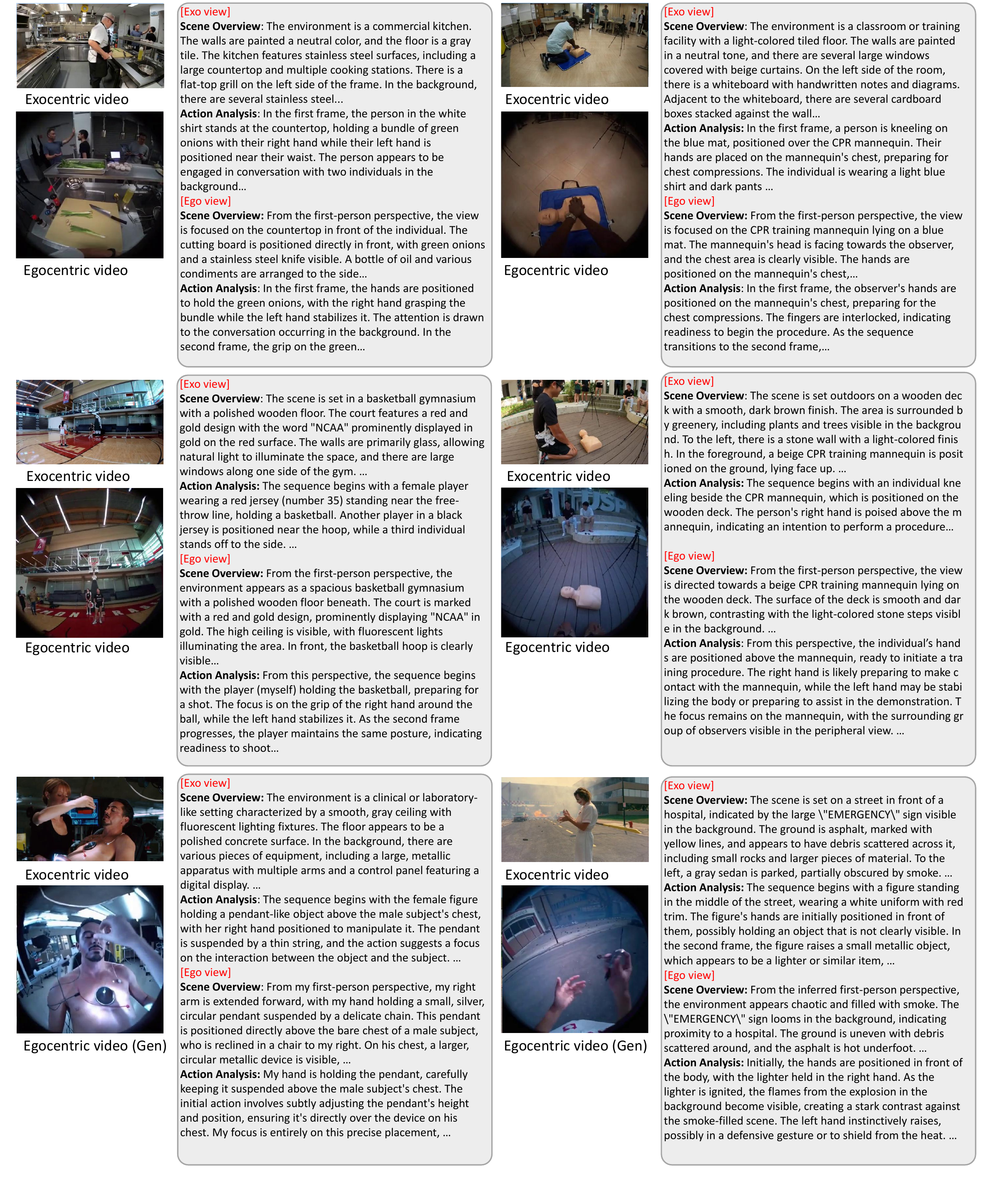}
    
   \caption{\textbf{Used Prompt Example}. This is the input text prompt for our model. Since the exocentric views were width-wise concatenated, the prompt describes both the exocentric and egocentric views.} 
   \label{fig:A_prompt}
\end{figure*}

Since our method builds on the pretrained diffusion model~\cite{wan}, text prompts are required to condition the model.
We generate these text prompts using a vision–language model (GPT-4o).
The system prompt used for generating these descriptions is provided in~\cref{tab:A_prompt_template}, and examples of the resulting text prompts can be found in~\cref{fig:A_prompt}.

\begin{table*}[t]
\centering
\resizebox{0.97\textwidth}{!}{
\begin{tabular}{l|ccccccccccccc}

\toprule
& \multicolumn{4}{c}{Image Criteria} & \multicolumn{3}{c}{Object Criteria} & \multicolumn{3}{c}{Video Criteria} \\
\cmidrule(lr){2-5} \cmidrule(lr){6-8} \cmidrule(lr){9-12}
Method & PSNR $\uparrow$ & SSIM $\uparrow$ & LIPIS  $\downarrow$ & CLIP-I $\uparrow$ & \begin{tabular}[c]{@{}c@{}}Location \\ Error \end{tabular} $\downarrow$ & IoU $\uparrow$ & \begin{tabular}[c]{@{}c@{}}Contour\\ Accuracy\end{tabular} $\uparrow$ & FVD $\downarrow$ & \begin{tabular}[c]{@{}c@{}} Temporal \\ Flickering \end{tabular} $\uparrow$ & \begin{tabular}[c]{@{}c@{}}Motion \\ Smoothness \end{tabular} $\uparrow$  & \begin{tabular}[c]{@{}c@{}}Dynamic \\ Degree \end{tabular} $\uparrow$\\
\drule
EgoX (Ours)                     & \textbf{16.05}    & \textbf{0.556}    & \textbf{0.498}    & 0.896             & \textbf{61.81}     & \textbf{0.363}    & \textbf{0.546}    & \textbf{184.47}& \textbf{0.977}& \textbf{0.989}& \textbf{0.974}   \\
w/o GGA                         & 14.77             & 0.539             & 0.530             & \textbf{0.897}    & 64.30              & 0.326             & 0.538              & 254.08         & 0.969        & 0.987    & 0.877\\

\midrule

Prior width, Exo Channel        & 13.83             & 0.471             & 0.594             & 0.736             & 83.08              & 0.213             & 0.501             & 274.14         & 0.964       & 0.986     & 0.915 \\
Prior width, Exo width          & 14.85             & 0.499             & 0.545             & 0.876             & 71.93              & 0.261             & 0.501             & 242.83         & 0.953       & 0.982     & 0.910 \\

\midrule
GGA only for inference          & 15.23             & 0.540             & 0.521             & 0.895             & 64.34              & 0.324             & 0.540             & 193.82         & 0.967       & 0.985     & 0.899 \\

\bottomrule

\end{tabular}
}
\caption{
\textbf{Additional Ablation Results.} The results from the conditioning strategy ablation and the GGA Training ablation are shown. These comparisons confirm that our integrated approach achieves the highest performance across all evaluated metrics.
\textbf{Best} results are highlighted in bold.
}
\label{tab:Appen_fusion_ablation}
\end{table*}

\section{In-depth Ablation Study}
\label{Appen:ablation}

\begin{figure}[t]
  \centering
   \includegraphics[width=1.0\linewidth]{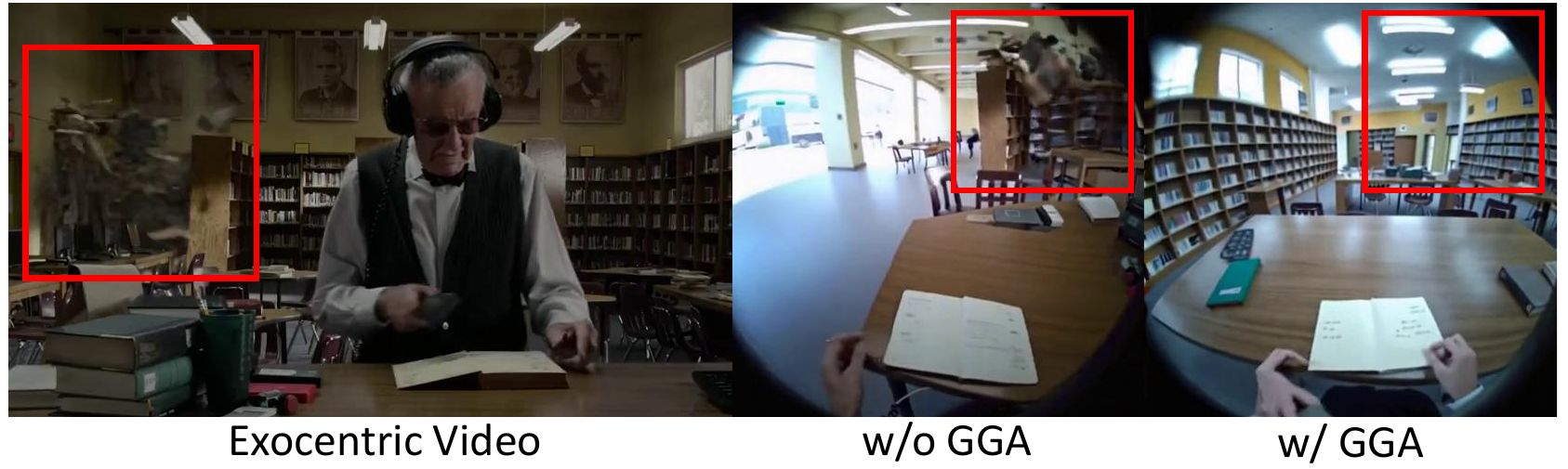}
    
   \caption{\textbf{GGA benefits example.} Without GGA, events occurring outside the visible region are attended to, leading to the generation of unwanted events in the ego view. With GGA, the model effectively focuses only on the visible region, thereby preventing the generation of these unwanted events.}
   \label{fig:A_GGA_benefit}
\end{figure}

\subsection{Ablation on Unseen Scene}
To further evaluate the generalization capability of each component, we additionally conduct ablation experiments on unseen scenes.
As shown in ~\cref{tab:Appen_exp_ablation}, the overall trends closely follow those observed in the seen-scene setting: removing any single component leads to noticeable degradation in geometric consistency, fidelity, or temporal coherence.
These results confirm that all three components, geometry-guided attention, the egocentric prior, and the clean latent strategy, are essential for achieving coherent, high-fidelity egocentric video generation, even in challenging unseen environments.

\subsection{Point cloud rendering}

To construct accurate egocentric prior frames, we employ monocular depth estimation~\cite{moge2} combined with depth alignment from ViPE~\cite{vipe}.
To validate the importance of depth alignment, we compare point cloud rendering with and without the alignment module.
As shown in ~\cref{fig:A_vda}, without depth alignment, monocular depth predictions exhibit frame-wise scale inconsistencies, causing even static background regions to shift across frames.
Although the ego camera remains fixed, misaligned depth introduces artificial camera motion, which can confuse the generative model and degrade viewpoint consistency.
In contrast, applying depth alignment corrects these temporal inconsistencies by ensuring that the depth scale is coherent across frames.
As a result, the rendered point clouds remain stable over time, providing a reliable egocentric prior for downstream video generation.

\subsection{Conditioning Strategy Ablation}


We evaluate how different conditioning strategies affect model performance by altering how the exocentric latent and the egocentric prior latent are combined.
Conceptually, the exocentric view, whose spatial alignment with the egocentric target is not pixel-consistent and requires implicit warping, should be conditioned in a way that preserves its global spatial structure, making width-wise concatenation a natural choice.
Conversely, the egocentric prior provides pixel-aligned viewpoint information, so channel-wise concatenation is better suited for injecting this fine-grained correspondence into the model.

To validate this intuition, we experiment with alternative fusion layouts.
One variant reverses the two conditioning directions, applying channel-wise concatenation to the exocentric latent and width-wise concatenation to the egocentric prior.
Another variant concatenates both inputs width-wise.
We do not test the setting where both inputs are concatenated channel-wise, as this requires adding extra network modules such as zero-convs.
When both inputs are concatenated width-wise, their combined latent becomes too large to fit in memory. Therefore, we resize the fused tensor to match the original exocentric latent shape.
Additionally, when the exocentric view is concatenated channel-wise, geometry-guided attention cannot operate because the spatial structure needed for warping is lost, so this variant is evaluated without GGA.

As shown in~\cref{tab:Appen_fusion_ablation} and~\cref{fig:A_add_ablation}, across all comparisons, our proposed conditioning strategy consistently delivers the strongest results.
When the exocentric latent is fused channel-wise, the model fails to learn the necessary warping behavior and cannot properly utilize the exocentric conditioning.
Similarly, width-wise concatenation of both latents diminishes the influence of the pixel-aligned prior and leads to quality degradation caused by confusion between global and local information.
In contrast, our design, width-wise concatenation for exocentric latents and channel, wise fusion for egocentric priors, achieves the best geometric alignment, the most reliable conditioning behavior, and the highest visual quality.

\begin{table}[t]
\centering
\setlength{\tabcolsep}{2pt}
\begin{tabular}{@{}l|cccc}
\toprule
Method                   & Ours             & -GGA             & -Ego Prior        & -Clean Latent         \\ \midrule
Runtime                  & $\sim$ 10.5 min    & $\sim$ 6.5 min     & $\sim$ 6.5   min    & $\sim$ 6.5   min        \\ \bottomrule
\end{tabular}
\caption{\textbf{Comparison of runtime for each component.} Runtime for each component was measured on an NVIDIA H200 GPU}
\label{tab:computational_cost}
\end{table}

\section{Additional Results}
\label{Appen:results}

\subsection{GGA Training Ablation}

\begin{figure*}[t]
  \centering
   \includegraphics[width=0.9\textwidth]{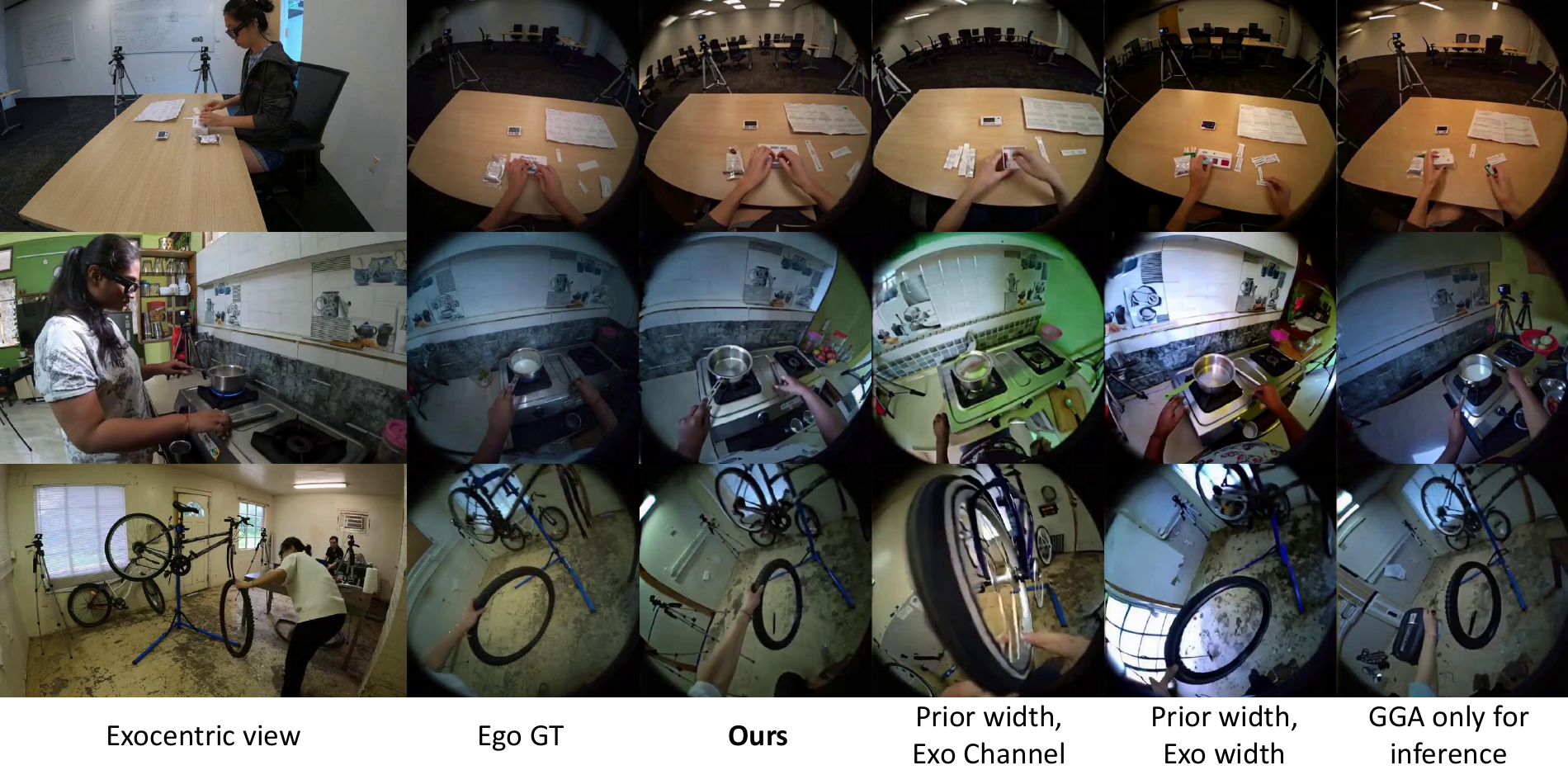}
   \caption{
   \textbf{Additional ablation qualitative comparison.}
   The qualitative results from the conditioning strategy ablation and the GGA Training ablation are shown.
   Model variants show limitations in geometric fidelity and detail reproduction, whereas our model consistently demonstrates the highest quality output.
   }
   \label{fig:A_add_ablation}
 \vspace{-\baselineskip}
\end{figure*}

To understand the role of geometry-guided attention (GGA) during learning, we compare two settings: applying GGA only at inference time versus applying it during both training and inference.
Because GGA serves as a guidance mechanism rather than a learnable module, one might expect it to be sufficient as an inference-only operation. However, when GGA is introduced solely at inference time, the model encounters an attention distribution it has never been trained to interpret. As shown in ~\cref{tab:Appen_fusion_ablation} and~\cref{fig:A_add_ablation}, this mismatch leads to a noticeable drop in visual fidelity and weaker geometric alignment.
In contrast, enabling GGA during training allows the model to learn attention patterns that naturally incorporate geometric bias. As a result, the model produces sharper details, more stable reconstructions, and significantly more accurate geometry during egocentric generation.

\subsection{Runtime}

We measure runtime based on the denoising stage, which is the most computationally intensive component of our pipeline.
Generating the egocentric prior takes less than 10 seconds and represents only a small fraction of the total processing time.
To quantify the overhead of each component, we evaluate variants of our system that disable individual modules.

GGA introduces a moderate overhead due to the additional attention bias computation required at every attention layer.
However, this cost is deemed highly reasonable and necessary for the significant overall performance improvements observed both qualitatively and quantitatively, particularly in areas like geometry and appearance. 
Crucially, the GGA provides essential guidance to the model. As illustrated in ~\cref{fig:A_GGA_benefit}, when generating the ego-view, the model without GGA may inadvertently attend to events occurring outside the visible region. This leads to the generation of unwanted events within the final ego-view. 
In contrast, GGA effectively guides the attention mechanism not to attend to these irrelevant regions, thereby preventing the generation of these undesirable events. This critical ability to ensure clean, accurate, and relevant ego-view generation makes the additional computational cost of GGA a worthwhile and necessary investment.
Using the egocentric prior incurs a similar cost to the difference between an image-to-video and text-to-video diffusion model, as it increases the input conditioning dimensionality without modifying the model architecture.
The clean latent strategy, however, adds no computational overhead, since it only modifies the noise scheduling during denoising without adding extra operations.

\subsection{User Study}
\begin{figure}[t]
  \centering
   \includegraphics[width=1.0\linewidth]{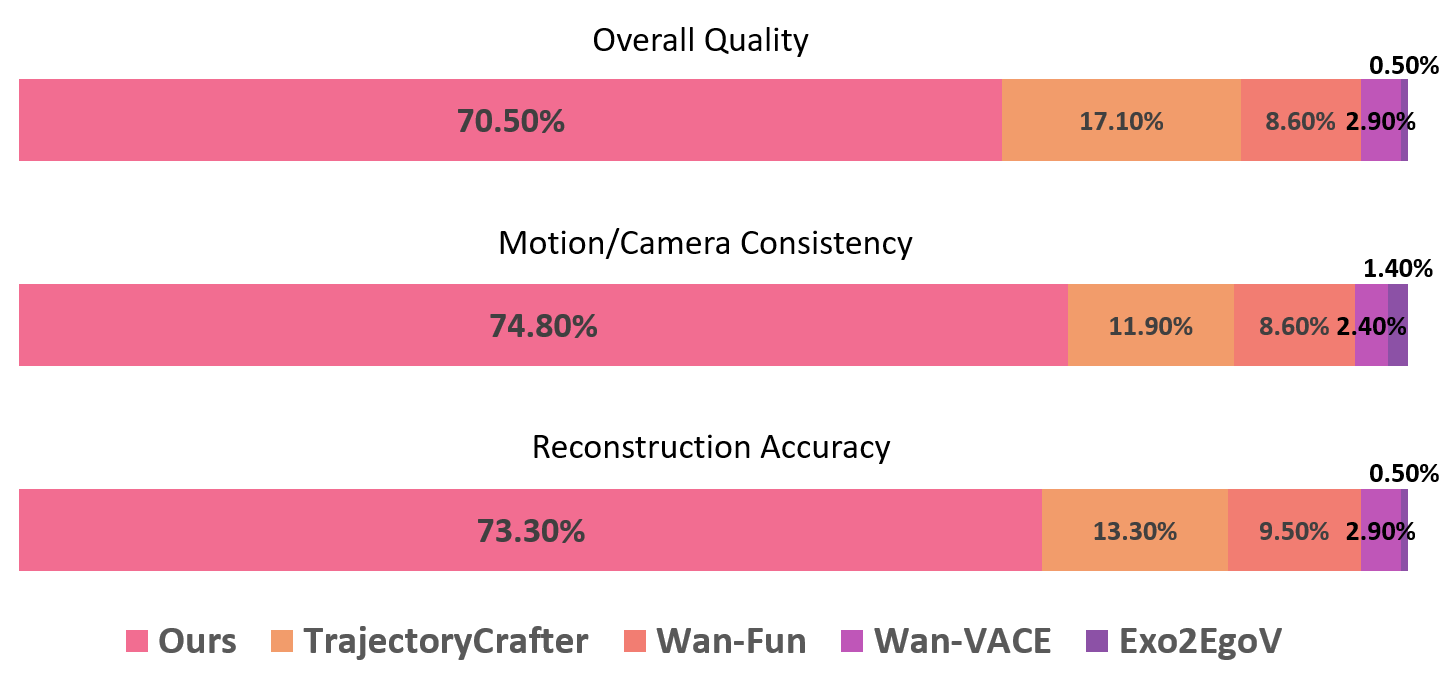}
    
   \caption{\textbf{User study results.} Our method received the highest number of selections across all questions, significantly outperforming all baselines.}
   \label{fig:A_user_study}
\end{figure}

To further evaluate the generalization capability of our method, we conducted a user study involving 20 unseen-scene videos and 10 in-the-wild videos.
A total of 19 participants were asked to choose the best video among the five methods, our method and four baselines, for each of the following criteria:
\begin{itemize}[leftmargin=1.2em]
    \item \textbf{Reconstruction Accuracy} Which result best preserves the content visible in the exocentric video?

    \item \textbf{Motion/Camera Consistency} Which result best follows the motion and camera trajectory observed in the exocentric view?

    \item \textbf{Overall Quality} Which result provides the highest overall egocentric video quality?
\end{itemize}
As shown in~\cref{fig:A_user_study}, our method received the highest number of selections across all questions, significantly outperforming all baselines.
These results demonstrate that our approach not only reconstructs view-relevant content more faithfully but also generalizes effectively to challenging unseen and in-the-wild scenarios.

\begin{figure}[t]
  \centering
   \includegraphics[width=1.0\linewidth]{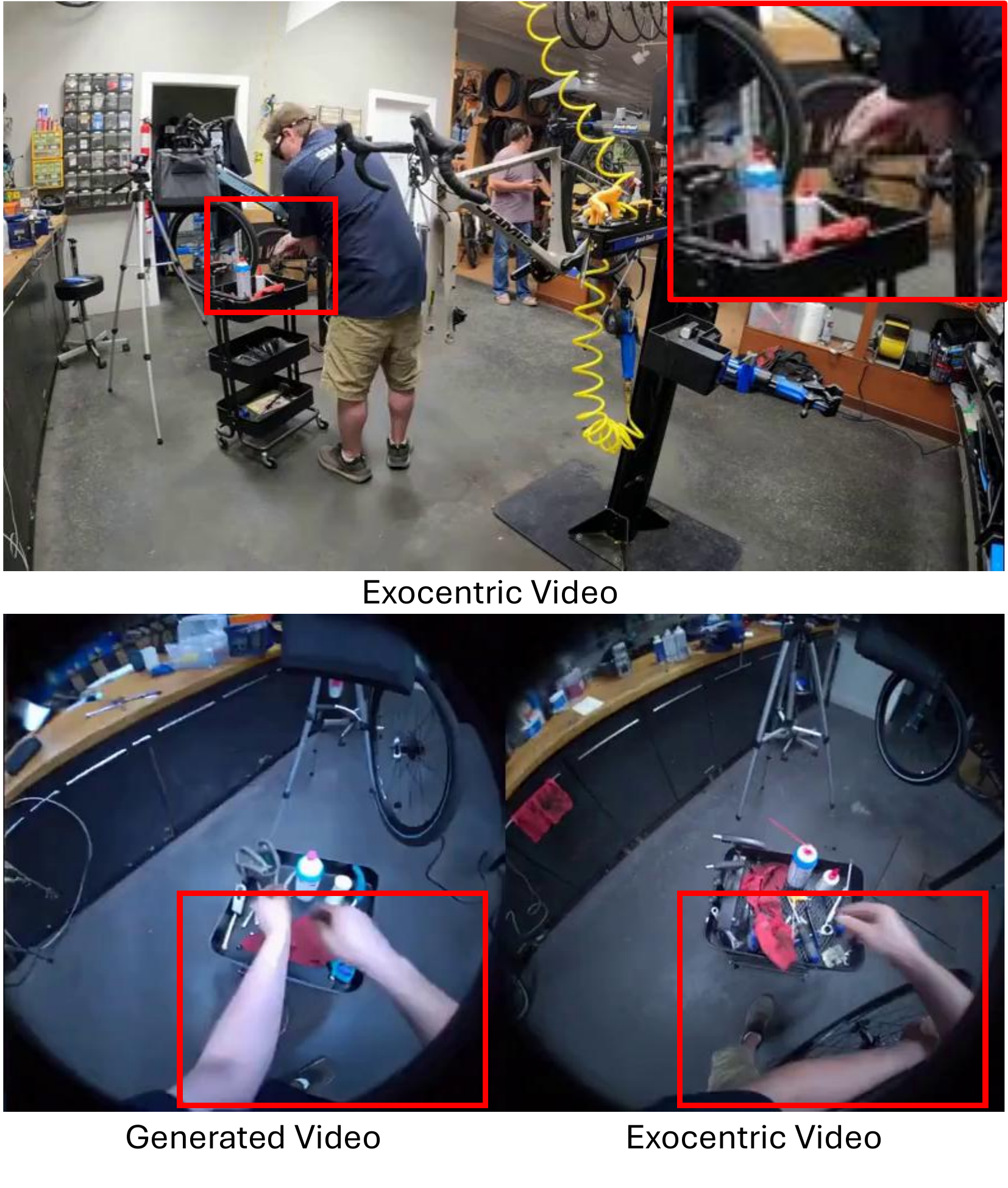}
    
   \caption{\textbf{Failure case due to task ambiguity.} The model's action misinterpretation occurs when it focuses on a small, subtle cue. This is not strictly a model failure, but rather a limitation imposed by the task's high ambiguity, where even a human observer might struggle to correctly infer the action based on such sparse visual evidence.}
   \label{fig:A_failure_case}
\end{figure}

\subsection{Additional qualitative results}
\begin{figure*}[t]
  \centering
   \includegraphics[width=0.9\textwidth]{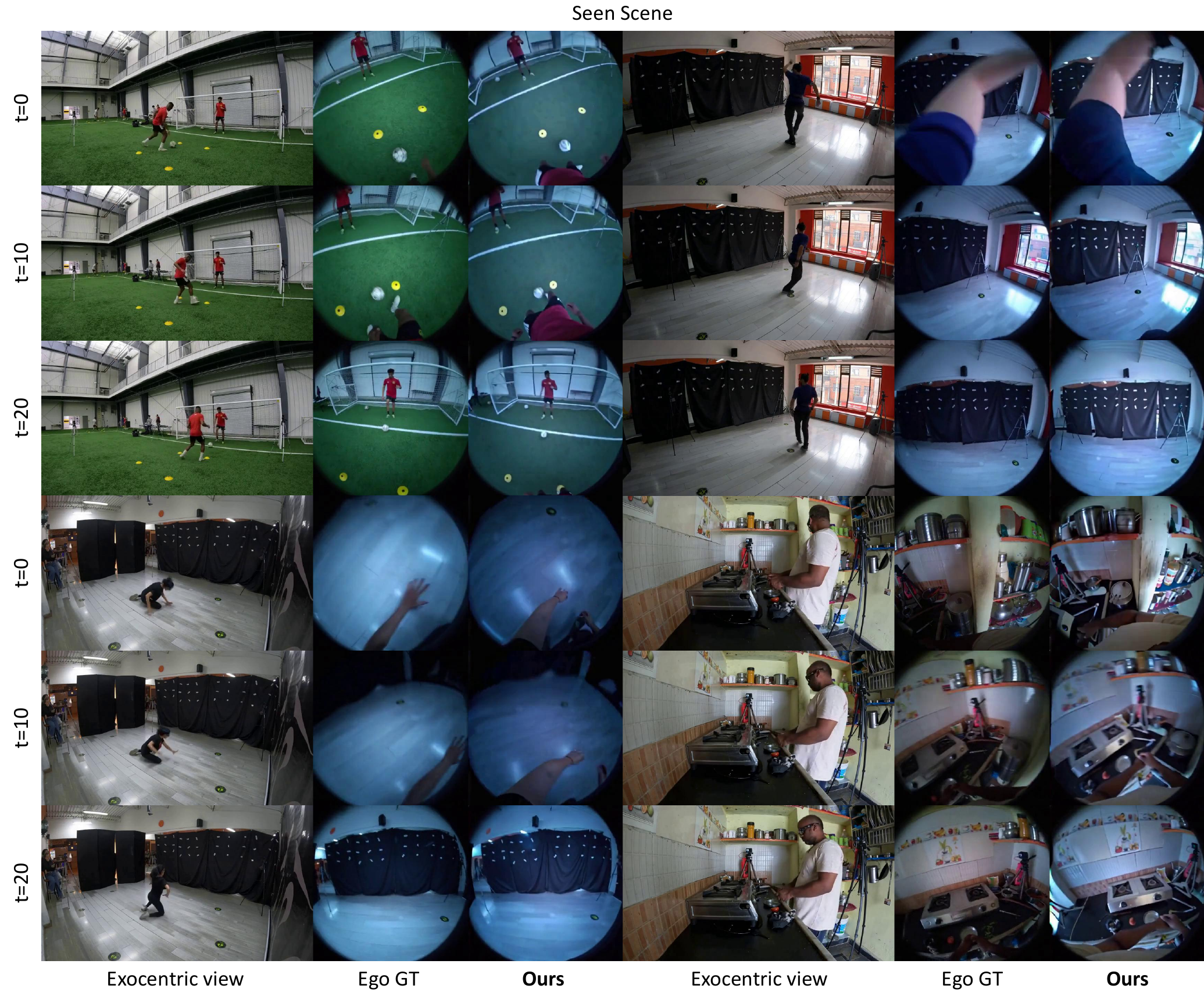}
   \caption{\textbf{Qualitative results for time sequence.} Our model accurately and seamlessly generates the entire time sequence.}
   \label{fig:A_add_quali}
 \vspace{-\baselineskip}
\end{figure*}

\begin{figure*}[t]
  \centering
   \includegraphics[width=0.9\textwidth]{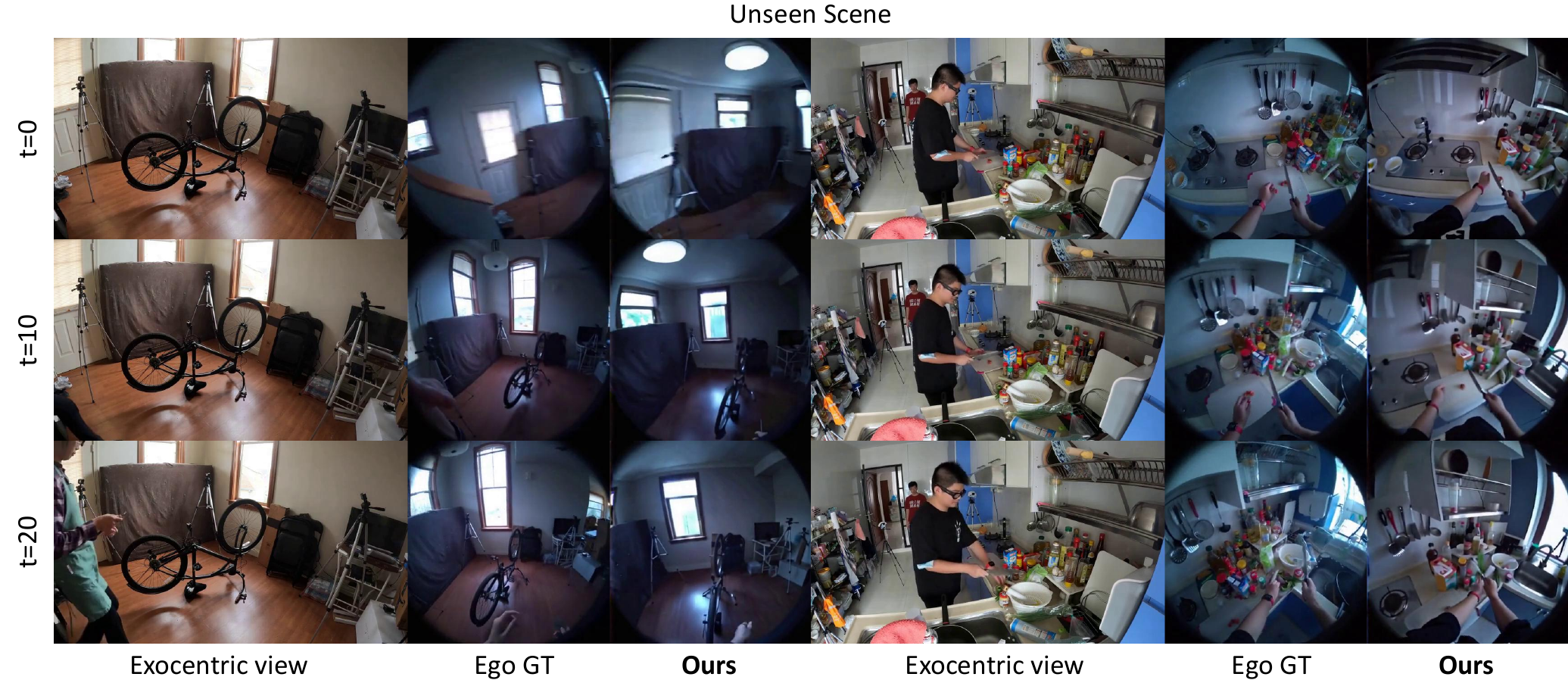}
   \caption{\textbf{Qualitative results for time sequence.} Our model accurately and seamlessly generates the entire time sequence.}
   \label{fig:A_add_quali_unseen}
 \vspace{-\baselineskip}
\end{figure*}

\begin{figure*}[t]
  \centering
   \includegraphics[width=0.9\textwidth]{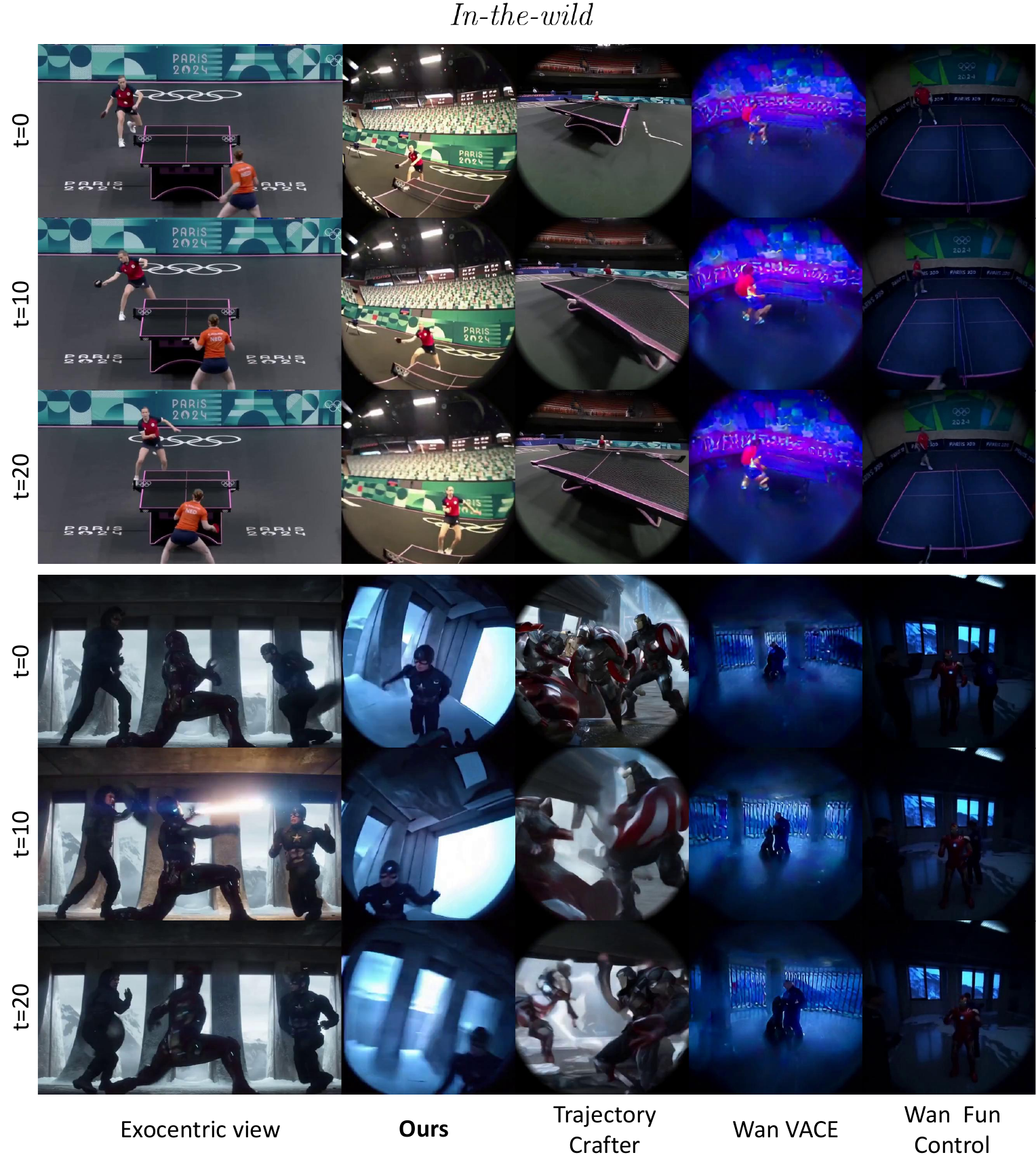}
   \caption{\textbf{Qualitative comparison with in-the-wild example.} Our model generates the entire time sequence accurately and seamlessly, even on challenging in-the-wild examples. Baselines struggle to maintain visual quality and accurate camera movements across all frames. }
   \label{fig:A_add_itw}
\end{figure*}

\begin{figure*}[t]
  \centering
   \includegraphics[width=1.0\textwidth]{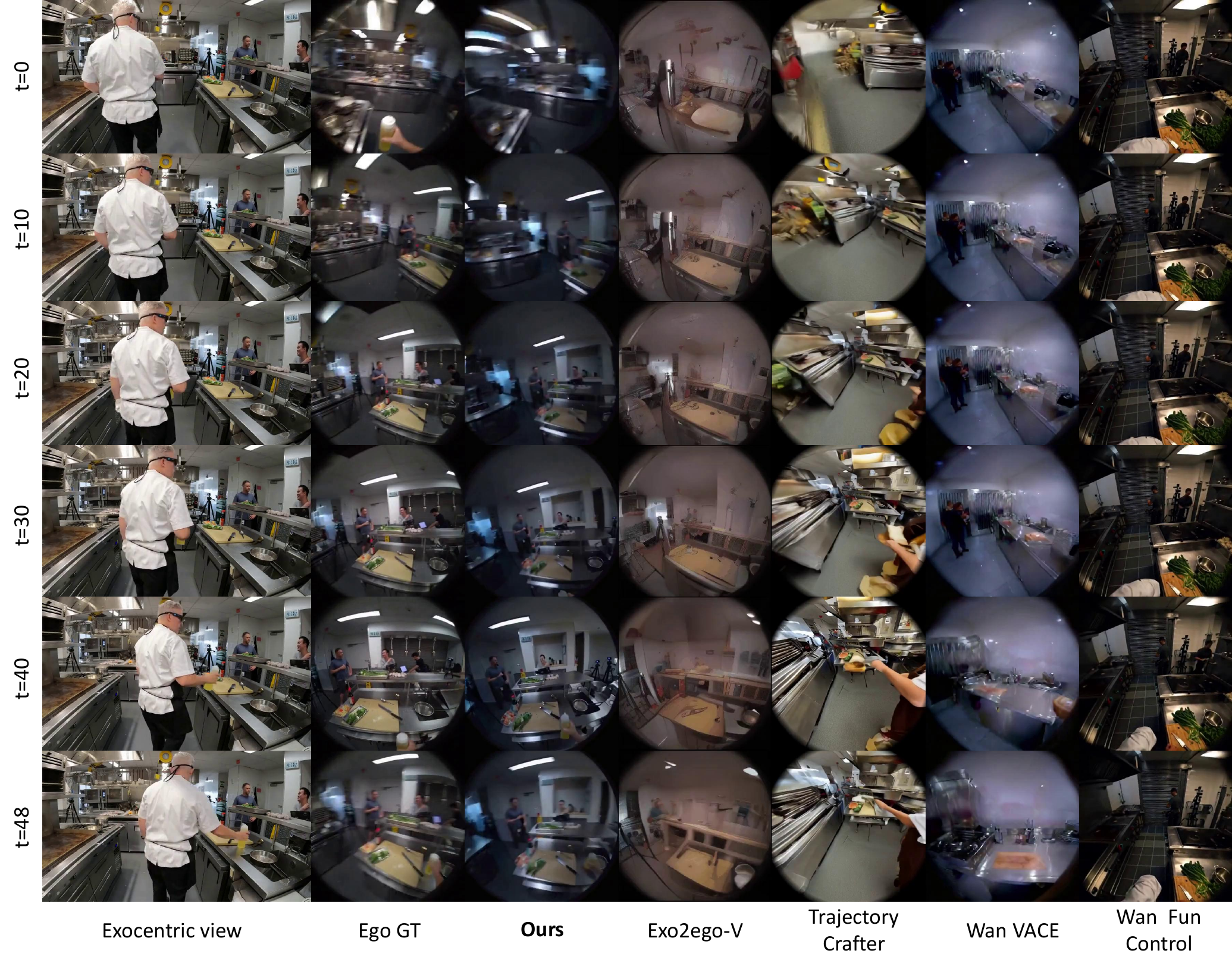}
   \caption{\textbf{Qualitative comparison for time sequence.} Our model accurately and seamlessly generates the entire time sequence. In contrast, other baselines struggle with maintaining high visual quality and generating accurate camera movements across all frames. }
   \label{fig:A_time_seq}
\end{figure*}

We include time-axis visualizations in~\cref{fig:A_add_quali,fig:A_add_quali_unseen,fig:A_time_seq}, which allow a clearer examination of temporal dynamics and overall video consistency.
Consistent with the quantitative metrics, our method produces natural, high-fidelity egocentric videos with accurate geometry and stable motion.
In contrast, Wan VACE often generates overly static videos with minimal dynamics, while other baselines either fail to properly incorporate the exocentric conditioning or exhibit noticeable artifacts and distortions.
We also include additional in-the-wild examples for time-axis visualizations in ~\cref{fig:A_add_itw}.

\subsection{Failure example}
Although our method performs robustly across diverse scenes, challenging real-world scenarios from datasets such as EgoExo4D~\cite{ego-exo4d} can still lead to occasional failure cases. These scenes often involve subjects facing away from the camera, rapid or complex body movements, or low-resolution details, making accurate cross-view reasoning extremely difficult.
As illustrated in~\cref{fig:A_failure_case}, when an exocentric frame contains ambiguous actions, such as a person bending one arm while the other arm is partially occluded, the model may misinterpret the configuration and generate an egocentric view with both arms extended. Such failure cases arise from inherent ambiguities in the exocentric input and the extreme viewpoint transformation required by the task.

\begin{table*}[!t]
\centering
\renewcommand{\arraystretch}{1.2}
\begin{tabular}{|p{\linewidth}|}
\hline
\textbf{System Prompt to obtain exo and egocentric video prompt} \\  
\hline
    You are a hyper-realistic scene reconstruction AI. Your task is to analyze a sequence of video frames 
    provided in chronological order and produce a comprehensive, two-part analysis: a static scene overview 
    followed by a dynamic, frame-by-frame action breakdown. Your guiding principle is \textbf{strict objectivity}.

    --- MISSION PROTOCOL ---

    \textbf{Phase 1: Scene Establishment}
    
    First, analyze all provided frames to establish a detailed, static description of the physical environment.
    Detail the surfaces (walls, floors), furniture, and all unmoving background items. This is your 'establishing shot'.

    \textbf{Phase 2: Action Transition Analysis}
    
    After establishing the scene, provide a detailed description of the action progression and transitions observed across the sequence. 
    Focus on how actions evolve, change, and flow from one moment to the next, maintaining awareness of the overall context established in Phase 1.

    --- CRITICAL DIRECTIVES ---

    \textbf{1. Exhaustive Object Inventory: THIS IS YOUR MOST IMPORTANT TASK.}
    
    You must meticulously identify and catalog EVERY visible item.
    
    - \textbf{NO GENERIC TERMS}: Do not use vague words like 'tool', 'box', 'utensil', or 'device'.
    
    - \textbf{BE SPECIFIC}: Use precise names (e.g., 'smartphone', 'coffee mug', 'wooden spoon', 'cutting board', 'refrigerator', 'laptop computer', 'ceramic bowl', 'stainless steel knife').
    
    - \textbf{DESCRIBE PROPERTIES}: Include colors, materials, textures, and positions (e.g., 'a blue ceramic mug on a granite countertop').

    \textbf{2. Focus on Hand-Object Interaction: THE ACTION'S CORE.}
    
    - For the `[Exo view]`, your primary narrative focus MUST be the person's hands.** Describe their precise posture, movement, and interaction with objects (e.g., 'the person's right hand grasps the knife handle,' 'the left hand's fingertips stabilize the tomato').
    
    - Every action description should revolve around what the hands are doing.

    \textbf{3. Strict Objectivity: DESCRIBE, DO NOT INTERPRET.}
    
    - \textbf{AVOID JUDGMENT}: Do not use subjective or abstract adjectives (e.g., AVOID 'modern', 'beautiful', 'cluttered', 'well-lit'). Describe only physical, measurable attributes.

    \textbf{4. Transition-Focused Analysis}
    
    - Analyze the sequence as a continuous flow of actions
    
    - Describe how movements and interactions transition and evolve
    
    - Focus on the progression and changes rather than individual frame descriptions
    
    - Maintain narrative continuity throughout the sequence

    --- OUTPUT STRUCTURE ---

    You MUST follow this exact two-block format:

    [Exo view]
    
    \textbf{Scene Overview:}
    Detailed description of the static background environment from the third-person perspective. List all background objects.

    \textbf{Action Analysis:}
    Describe the progression of actions and transitions observed throughout the sequence. Focus on how movements evolve, interactions change, and the flow of activities from beginning to end. Describe the continuous narrative of what is happening.

    [Ego view]
    
    \textbf{Scene Overview:}
    Detailed description of the static background environment from the first-person perspective. List all background objects.

    \textbf{Action Analysis:}
    Describe the progression of actions and transitions observed throughout the sequence from the first-person perspective. Focus on how movements evolve, interactions change, and the flow of activities from beginning to end. Describe the continuous narrative of what is happening from the ego viewpoint.

    \{image\}  \\

\hline
\end{tabular}
\caption{
\textbf{System Prompt for VLM.} This is the system prompt used to generate the input text prompt for our model. Since the exocentric views were width-wise concatenated, the prompt describes both the exocentric and egocentric views. 
}
\label{tab:A_prompt_template}
\end{table*}

\end{document}